\documentclass{article}
\PassOptionsToPackage{numbers, compress}{natbib}

\usepackage[final]{neurips_2022}

\usepackage[utf8]{inputenc} 
\usepackage[T1]{fontenc}    
\usepackage{hyperref}       
\usepackage{url}            
\usepackage{booktabs}       
\usepackage{amsfonts}       
\usepackage{nicefrac}       
\usepackage{microtype}      
\usepackage{xcolor}         
\usepackage{graphicx}
\usepackage{amsmath}
\usepackage{bm}
\usepackage{makecell}
\usepackage{tabularx}
\usepackage{multirow}
\usepackage{amssymb}
\usepackage{amsmath}
\usepackage{tabularx}
\usepackage{bm}
\usepackage[ruled,linesnumbered]{algorithm2e}
\usepackage{algorithmic}
\usepackage{xcolor}
\usepackage{listings}
\newcolumntype{Y}{>{\centering\arraybackslash}X}

\usepackage{soul} %
\usepackage{color, xcolor} %
\soulregister\cite7 
\soulregister\citep7 
\soulregister\citet7 
\soulregister\ref7 
\soulregister\pageref7 
\def\degree{${}^{\circ}$}

\newcommand{\ie}{\textit{i.e.}}
\newcommand{\eg}{\textit{e.g.}}

\title{FreGAN: Exploiting Frequency Components for Training GANs under Limited Data}

\author{
  Mengping Yang$^\dagger$$^\ddagger$ \quad
  Zhe Wang$^\dagger$$^\ddagger$\thanks{Corresponding author} \quad
  Ziqiu Chi$^\dagger$$^\ddagger$ \quad
  Yanbing Zhang$^\dagger$$^\ddagger$ \\
  $^\dagger$Department of Computer Science and Engineering, ECUST, China \\
  $^\ddagger$Key Laboratory of Smart Manufacturing in Energy Chemical Process, ECUST, China \\
  \texttt{wangzhe@ecust.edu.cn} \\
  \texttt{mengpingyang@mail.ecust.edu.cn} \\
}

\begin{document}

\maketitle
%
\begin{abstract}
\label{sec:abstract}
Training GANs under limited data often leads to discriminator overfitting and memorization issues, causing divergent training. Existing approaches mitigate the overfitting by employing data augmentations, model regularization, or attention mechanisms. However, they ignore the frequency bias of GANs and take poor consideration towards frequency information, especially high-frequency signals that contain rich details. To fully utilize the frequency information of limited data, this paper proposes FreGAN, which raises the model's frequency awareness and draws more attention to producing high-frequency signals, facilitating high-quality generation. In addition to exploiting both real and generated images' frequency information, we also involve the frequency signals of real images as a self-supervised constraint, which alleviates the GAN disequilibrium and encourages the generator to {synthesize} adequate rather than arbitrary frequency signals. Extensive results demonstrate the superiority and effectiveness of our FreGAN in ameliorating generation quality in the low-data regime (especially when training data is less than 100). Besides, FreGAN can be seamlessly applied to existing regularization and attention mechanism models to further boost the performance.
\footnote{Our codes are available at \url{https://github.com/kobeshegu/FreGAN_NeurIPS2022}.}

\end{abstract}


\section{Introduction}
\label{sec:introduction}
Generative adversarial networks (GANs)~\cite{goodfellow2014generative} have shown impressive achievements in synthesising plausible and photorealistic visual objects, such as image~\cite{karras2019style}~\cite{Karras2021} and video~\cite{wang2019event} generation, image inpainting~\cite{liu2021pd}, image translation~\cite{tov2021designing} and so on.
However, a prerequisite of such success is sufficient training data, which impedes applications of GANs in areas where only dozens of data are available or where it is challenging to collect massive data due to geographical, spatial, temporal, or privacy reasons.
Thus developing data-efficient GANs that can generate plausible images under limited data, without compromising the quality, is necessary and meaningful.

\begin{figure}
	\centering
	\vspace{-0.2cm}
	\includegraphics[width=\linewidth]{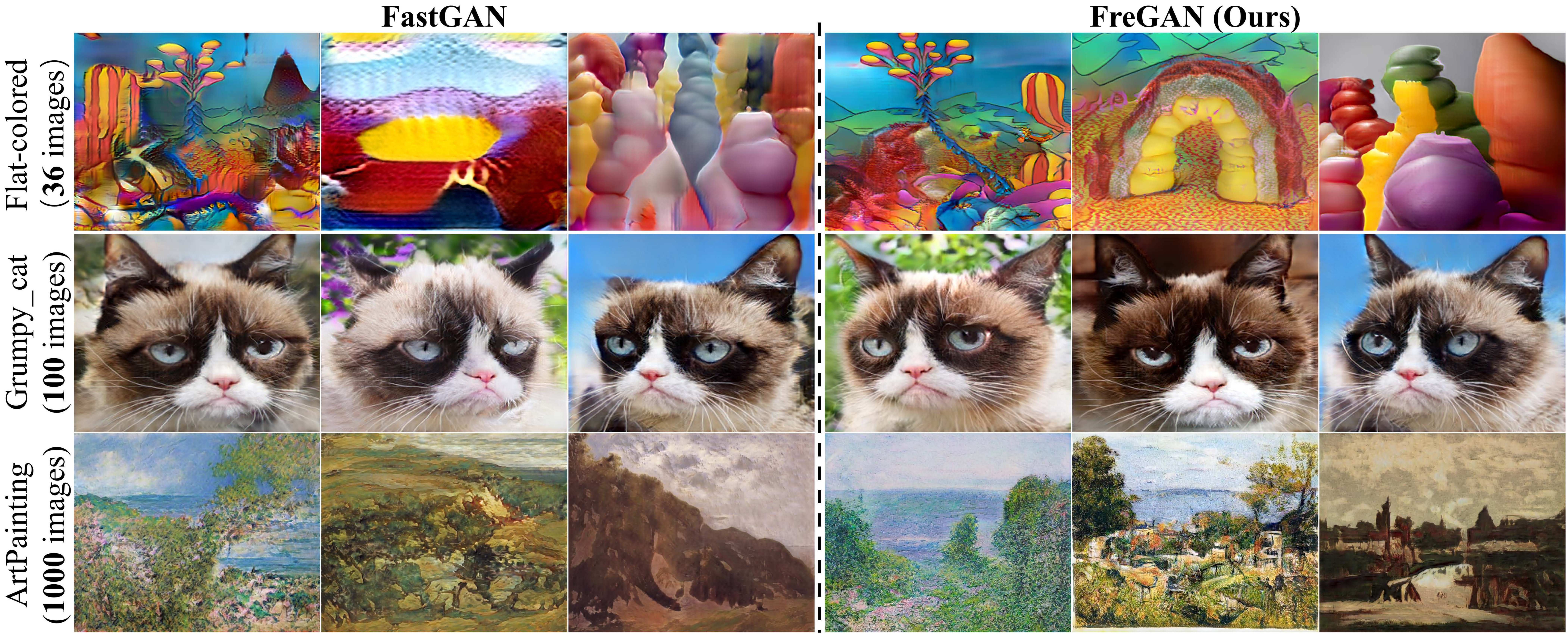}
	\vspace{-0.4cm}
	\caption{\textbf{Images generated by FastGAN~\cite{liu2021towards} and our FreGAN given limited training data.} The details of FastGAN deteriorate while our FreGAN effectively ameliorates the {synthesize} quality by raising the model's frequency awareness, producing plausible images with better fine details.}
	\label{fig:introvis}
	\vspace{-6mm}
\end{figure}

Training GANs under limited data often leads to overfitting and instability issues~\cite{karras2020training}~\cite{jiang2021deceive}.
Specifically, when the discriminator (D) overfits to the limited training data, it simply remembers the input real images and classifies others as fake images, thus providing meaningless feedback to the generator (G), leading to divergent training and poor-quality generation.
Ameliorating the {synthesize} quality under limited data is still an unexplored problem.
Recent approaches for this problem include enlarging the training set with different data augmentations~\cite{zhao2020image}~\cite{tran2021on}~\cite{DiffAug}~\cite{karras2020training}~\cite{jiang2021deceive}, regularizing the output of D with an  additional constraint~\cite{tseng2021regularizing},
and devising new network architectures~\cite{liu2021towards}.
However, existing methods are mainly developed from the perspective of data scale and model capacity, and they ignore a critical property of the data itself, \ie, frequency signals.
GANs have been demonstrated to have a spectral bias in fitting frequency signals~\cite{rahaman2019spectral}~\cite{tancik2020fourier}.
They preferentially fit low-frequency signals and tend to ignore high-frequency signals~\cite{xu2019frequency}, which encode fine details like vertical and horizontal edges~\cite{yoo2019photorealistic}~\cite{gao2021high}.
Missing them may lead to unrealistic image {synthesize} with unsatisfactory artifacts (see Fig.~\ref{fig:introvis}).
This paper proposes a frequency-aware model, termed as FreGAN, to raise the frequency awareness of G and D. By encouraging G to generate more reasonable and adequate high-frequency signals, our FreGAN ameliorates the {synthesize} quality under limited data, as shown in Fig.~\ref{fig:introvis}.

To fully exploit the frequency information of limited training data, we first decompose images into different frequency components via Haar wavelet transformation~\cite{daubechies1990wavelet}.
Unlike traditional wavelet transformation that is employed at the image level, we perform it on the intermediate features of both D and G.
We then employ {a} high-frequency discriminator (HFD) and frequency skip connection (FSC) to raise the frequency awareness of G and D, respectively.
However, G still has no explicit clue about what high-frequency signals it should {synthesize}, and D is overconfident in making real/fake decisions after seeing real and fake images.
Such an unbalanced competition motivates us to perform high-frequency alignment (HFA) to alleviate the information asymmetry between G and D.
Innovatively, we explicitly exploit the frequency signals of real images induced from D as a self-supervised constraint to guide G to leverage the frequency knowledge properly.
Besides, HFD and HFA are applied on multi-scale features to thoroughly excavate the frequency signals of limited data, mitigating frequency bias and loss of high-frequency information.

%
%

The primary contributions of this paper are three-fold:
1) we propose FreGAN to raise the model's frequency awareness, which successfully mines and exploits frequency information of limited data, and as a byproduct, FreGAN alleviates the unhealthy competition between G and D;
2) we demonstrate the compatibility of our model by combining our method with other techniques like regularization~\cite{tseng2021regularizing} and attention mechanism~\cite{li2021prototype};
3) we perform extensive experiments on various datasets with limited data, and our FreGAN achieves state-of-the-art performance on these datasets, indicating the effectiveness and superiority of our method for ameliorating {synthesize} quality, especially when training data is extremely limited.

%






\section{Related Work}
\label{sec:relatedwork}
\paragraph{Generative Adversarial Networks.}
Generative adversarial networks (GANs)~\cite{wang2021generative}~\cite{karras2018progressive}~\cite{WGANGP}, which target at generating plausible and realistic images, have made massive progress since the pioneering work~\cite{goodfellow2014generative}.
The capability of GANs enables various visual applications like image~\cite{karras2019style}~\cite{karras2020analyzing} and video generation~\cite{wang2019event}~\cite{tulyakov2018mocogan}, image inpainting~\cite{yu2021wavefill}~\cite{liu2021pd}, image manupulation~\cite{gao2021high}~\cite{tov2021designing} and super-resolution~\cite{wang2021urnet}, etc.
However, GANs are notoriously difficult to train as several issues like mode collapse and instability happen easily.
Numerous techniques have been proposed to stabilize training and improve the {synthesize} quality by designing new optimization objectives or network architectures~\cite{jabbar2021survey}.
WGAN~\cite{arjovsky2017wasserstein} and $f$-GAN~\cite{nowozin2016f} minimize the Wasserstein distance and the $f$-divergence of real and generated distribution instead of minimizing JS divergence in~\cite{goodfellow2014generative}.
BigGAN~\cite{donahue2019large} and StyleGAN series~\cite{karras2018progressive}~\cite{karras2019style}~\cite{karras2020analyzing}~\cite{karras2020training}~\cite{Karras2021} have made breakthrough progress in producing realistic images.
{However, the performance of these models deteriorates when given limited data}.

\paragraph{Training GANs under limited data.}
Improving the {synthesize} quality under limited data remains an underexplored problem, which has drawn extensive attention recently.
Insufficient training data leads to discriminator overfitting, thus degrading the quality of generated images.  
One straightforward way to address such data scarcity is to expand the training set with various augmentations.
In addition to employing conventional augmentation techniques~\cite{tran2021on}~\cite{zhao2020image} (e.g., flip, crop), ADA~\cite{karras2020analyzing} and DiffAug~\cite{DiffAug} propose adaptive and differentiable augmentation to enlarge the training data, respectively.
APA~\cite{jiang2021deceive} deceives D based on the degree of overfitting with an adaptive pseudo augmentation.
InsGen~\cite{yang2021data} involves instance discrimination as an auxiliary task to encourage D to distinguish every individual image, which improves the discriminative power of the discriminator.
Lecam~\cite{tseng2021regularizing} regularizes the output of the discriminator throughout the training process.
FastGAN~\cite{liu2021towards} employs a skip-layer channel-wise excitation module and a self-supervised discriminator to stabilize and accelerate the training.
The most recently MoCA~\cite{li2021prototype} improves few-shot image generation quality with a prototype memory with an attention mechanism.
Benefit from the significant progress of large-scale pre-trained visual recognition models, Vision-aided GAN~\cite{kumari2022ensembling} uses available off-the-shelf models to help the GAN training and ProjectedGAN~\cite{sauer2021projected} improve GANs by projecting generated and real images into pre-trained feature spaces.
Another category of methods transfer and reuse knowledge from models that are pre-trained on large-scale data, \ie, few-shot GAN adaptation~\cite{wang2018transferring}~\cite{li2020few}~\cite{wang2020minegan}~\cite{ojha2021fsgan}.
In this paper, we ameliorate the {synthesize} quality under limited data from the frequency domain perspective.
By raising the frequency awareness of GANs and providing more fine details to G, we facilitate photorealistic image generation.
Our work is complementary to previous model regularization and attention mechanism approaches, and our method promotes equilibrium between G and D.

\paragraph{Wavelet Transformation in GANs.}
Schwarz et al.~\cite{schwarz2021frequency} prove that GANs exhibit a frequency bias and resolving frequency artifacts is necessary for photorealistic image generation.
Consequently, GANs tend to ignore high-frequency signals as they are hard to generate, compromising the generation quality.
Wavelet transformation~\cite{daubechies1990wavelet}, which decomposes images into frequency components with different bands, has been wildly used in various applications of GANs, such as style transfer~\cite{chiu2022photowct2}~\cite{yoo2019photorealistic}, image inpainting~\cite{yu2021wavefill}, image editing~\cite{gao2021high}, etc.
HiFA~\cite{gao2021high} alleviates the generator's pressure of producing high-frequency signals by directly feeding high-frequency components to the generator.
WaveFill~\cite{yu2021wavefill} disentangles different frequency signals and explicitly fills the missing regions in each frequency band, achieving superior image inpainting.
Zhang et al.~\cite{zhang2022wavelet} propose wavelet knowledge distillation towards efficient image-to-image translation without a performance drop.
{SWAGAN~\cite{gal2021swagan} incorporates wavelet with the hierarchical training of StyleGAN2~\cite{karras2020analyzing} and performs wavelets at the image level. Our FreGAN is more flexible by directly decomposing intermediate features of the generator and the discriminator into the wavelet domain, and no additional down/up sampling are required to convert images to higher/lower resolution as in SWAGAN, which makes our method more efficient.}
Unlike existing methods that are performed on ample data, this paper addresses the more challenging few-shot generation problem.
In addition to raising the frequency awareness of the model, we also mitigate the unhealthy competition by lessening the frequency gap between G and D.


\section{Methodology}
\label{sec:method}

\begin{figure}
	\centering
	\vspace{-0.2cm}
	\includegraphics[width=\linewidth]{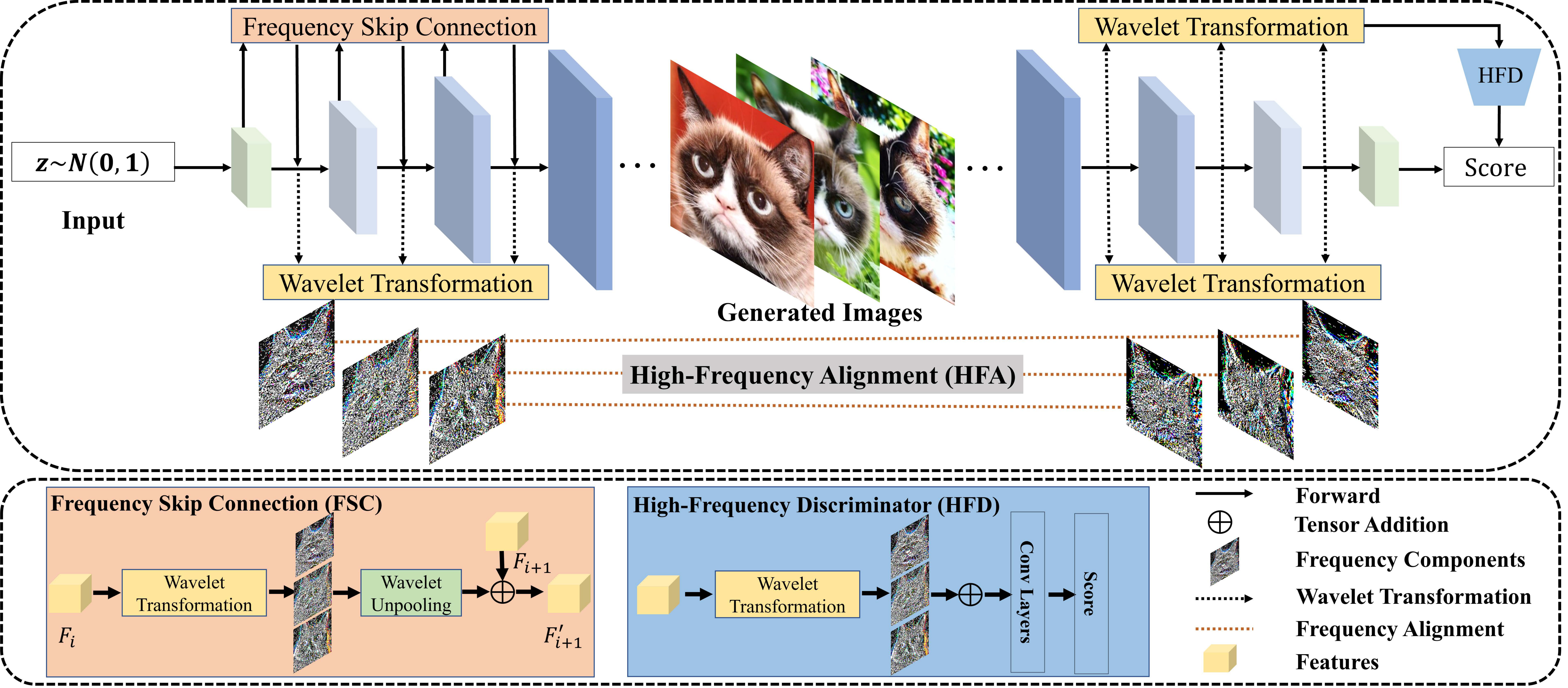}
	\vspace{-0.4cm}
	\caption{\textbf{The overall framework of our proposed FreGAN.} Composed of three key ingredients, \ie, frequency skip connection (FSC), high-frequency discriminator (HFD), and high-frequency alignment (HFA), our FreGAN raises the model's frequency awareness, facilitating high-quality image {synthesize} under limited data.}
	\label{fig:framework}
	\vspace{-4mm}
\end{figure}
The overall framework of our FreGAN is illustrated in Fig.~\ref{fig:framework}.
To formulate our method, we explicitly utilize wavelet transformation to decompose features into different frequency components.
We then employ high-frequency discriminator (HFD) and frequency skip connection (FSC) to raise the frequency awareness of G and D, respectively.
Moreover, we perform high-frequency alignment (HFA) to further guide G to {synthesize} adequate frequency signals.

\subsection{Wavelet Transformation}
\label{sec:wavelettransformation}

\begin{figure}
	\centering
	\includegraphics[width=\linewidth]{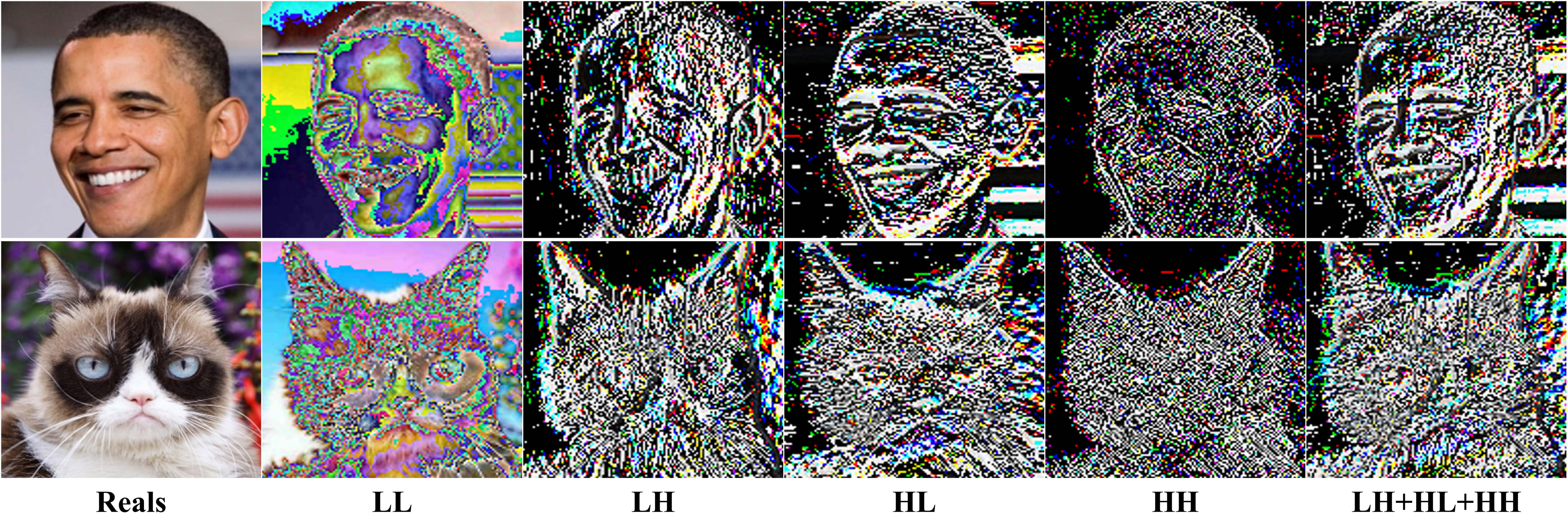}
	\vspace{-0.4cm}
	\caption{\textbf{Illustration of frequency components obtained from Haar wavelet transformation.} The low (L) pass filter captures images' overall textures and outlines, and the high (H) pass filter concentrates on details such as the background and edges.}
	\label{fig:frequency}
	\vspace{-6mm}
\end{figure}

To decompose images into different frequency components, we adopt a simple but effective wavelet transformation, \ie, Haar wavelet.
Haar wavelet consists of two mirror operations: wavelet pooling and wavelet unpooling.
The former converts images into the hl{wavelet} domain, and the latter inversely reconstructs frequency components into the spatial domain.
There are four kernels in wavelet pooling operation: ${LL^T, LH^T, HL^T, HH^T}$, where $L^{T}=\frac{1}{\sqrt{2}}[1,1]$,  $H^{T}=\frac{1}{\sqrt{2}}[-1,1]$, $L$ and $H$ denotes the low and high pass filters, respectively.
The low (L) pass filter captures the outline and surface of images, while the high (H) pass filter focuses on detailed information like the edges and delicate textures.
Fig.~\ref{fig:frequency} illustrates the obtained frequency components of given images via Haar wavelet.
We can observe that low-frequency component $LL$ contains the overall surface of images, while components that are decomposed by high pass filters, i.e., $LH, HL, HH$, contain more fine details.
Further, by summing up the three high-frequency components, we approximately obtain all details information of images, \eg, the eyes of the cat and the teeth of Obama.

\subsection{High-Frequency Discriminator}
\label{sec:HFD}
To raise the frequency awareness of D, we devise high-frequency discriminator (HFD).
HFD is responsible for distinguishing the real images from the generated images from the perspective of the frequency domain.
Formally, for the $i$-th layer in the discriminator, we adopt wavelet pooling on the intermediate features and obtain ${LL}^i_{D}, {LH}^i_{D}, {HL}^i_{D}, {HH}^i_{D}$, then we combine the three high-frequency components by tensor addition, \emph{i.e.}, $HF^i_{D} ={LH}^i_{D} + {HL}^i_{D} + {HH}^i_{D}$, which contains sufficient details of features.
By applying traditional convolution and downsampling operations following the original discriminator, we define the adversarial loss of our HFD as:
\begin{equation}
\mathcal{L}_{D}^{HF}=-\mathbb{E}_{{HF}_{\text{real}} \sim I_{\text {real}}}[\min (0,-1+D_H({HF}_{\text{real}}))]-\mathbb{E}_{{HF}_{\text{fake}} \sim G(z)}[\min (0,-1-D_H({HF}_{\text{fake}})]
\vspace{-2mm}
\end{equation}
\begin{equation}
\label{eq:Gloss}
\mathcal{L}_{G}^{HF}=\mathbb{E}_{HF_{\text{fake}}}[D_H(HF_{\text{fake}})]
\end{equation}
where $HF_{\text{real}}$ and $HF_{\text{fake}}$ are the high-frequency information of real and fake images, respectively. $D_H$ is the high-frequency discriminator.
Since the high-frequency information may be eschewed by D as the network goes deeper, we perform multi-scale HFD on multi-layers of the discriminator.
The multi-scale operation ensures fully mine and exploit the frequency information of limited data, which further improves D's frequency awareness.
Notably, being guided by the HFD with Eq.~\ref{eq:Gloss}, G is also optimized to produce rich high-frequency details.

\subsection{Frequency Skip Connection}
\label{sec:GSkip}
{The generator is capable of producing plausible frequency signals after employing HFD (see Tab.~\ref{tab:ablationmain})}.
However, as GANs fit frequency signals from low to high and the high-frequency signals may be ignored as the network goes deeper.
To prevent the loss of high-frequency information and further encourage the generator to produce rich details, we propose frequency skip connection (FSC).
Concretely, we utilize wavelet unpooling operation of the frequency components ${LL}^i_{G}, {LH}^i_{G}, {HL}^i_{G}, {HH}^i_{G}$ obtained from wavelet transformation on the features of G's $i$-th layer, which reconstructs the high-frequency representation to the original features.
Then we explicitly feed the reconstructed frequency representations to the next layer of G.
Formally,
\begin{equation}
{F}_{i+1}^{'}= {F}_{i+1} + Unpooling({LL}^i_{G}, {LH}^i_{G}, {HL}^i_{G}, {HH}^i_{G})
\end{equation}
where ${F}_{i}$ denotes the features of the $i$-th layer and $Unpooling$ is the wavelet unpooling operation.
${F}_{i+1}^{'}$ is the obtained features after FSC, which will be fed into the subsequent layer for further operation.
Such skip connection prevents loss of high-frequency information and maintains high-frequency details.

\subsection{High-Frequency Alignment}
\label{sec:FDA}
Adding HFD and FSC explicitly raises the frequency awareness of G, but G can only {synthesize} arbitrary frequency signals.
How G can utilize the frequency signals is still ambiguous, and D still dominates the competition since it learns discriminative knowledge from both real and generated images.
To balance the unhealthy competition between G and D, we propose high-frequency alignment (HFA), which involves high-frequency signals of real images induced from D as a regularizer to guide {G}, promoting G to {synthesize} more reasonable and realistic fine details.
Specifically, we extract the frequency representations of intermediate features of G at different layers.
For the $i$-th layer of G, we obtain frequency components ${LL}^i_{G}, {LH}^i_{G}, {HL}^i_{G}, {HH}^i_{G}$.
We ignore ${LL}^i_{G}$ and combine the three high-frequency components, \ie, $HF^i_{G} ={LH}^i_{G} + {HL}^i_{G} + {HH}^i_{F}$.
Then we use the high-frequency components of the discriminator $HF^i_D$ as a self-supervision constraint.
In addition to fool D, G is expected to minimize the distance of high-frequency information between the generated and real images.
The alignment loss is defined as:
\begin{equation}
\label{eq:HFA}
\mathcal{L}_{\text {align }}=\left\|HF_{D}-HF_{G} \right\|_{1}
\end{equation}
where $\|*\|_1$ denotes the $L_1$-norm.
Such alignment encourages G to synthesis frequency signals that approach real frequency signals, mitigating the unhealthy competition and facilitating generation quality.
To take full advantage of frequency signals of real images from D, we perform HFA on multi-scale features like HFD as shown in Fig.~\ref{fig:framework}.
The ablative experiment results in Sec.~\ref{sec:ablationstudy} demonstrate the rationality and effectiveness of employing HFA and HFD on multi-scale features.

\subsection{Optimization}
Following~\cite{liu2021towards}, we adopt the hinge version of adversarial loss to train our model.
\begin{equation}
\mathcal{L}_{D}=-\mathbb{E}_{x \sim I_{\text {real }}}[\min (0,-1+D(x))]-\mathbb{E}_{\hat{x} \sim G(z)}[\min (0,-1-D(\hat{x})]
\end{equation}
\vspace{-2mm}
\begin{equation}
\mathcal{L}_{G}=-\mathbb{E}_{z \sim \mathcal{N}}[D(G(z))]
\end{equation}
We also use the reconstruction loss~\cite{liu2021towards} to encourage the discriminator to extract more representative features.
\begin{equation}
\mathcal{L}_{\text {recons }}=\mathbb{E}_{\mathbf{f} \sim D_{\text {encode }}(x), x \sim I_{\text {real }}}[\|\mathcal{G}(\mathbf{f})-\mathcal{T}(x)\|]
\end{equation}
{where $\mathbf{f}$ is the intermediate features of D, $\mathcal{G}$ and $\mathcal{T}$ denote the processing on the features $\mathbf{f}$ and the input images $x$.}
In sum, our discriminator is optimized by $\mathcal{L}_{D}$, $\mathcal{L}_{\text {recons }}$, and $\mathcal{L}_{D}^{HF}$.
Our generator is optimized by $\mathcal{L}_{G}$, $\mathcal{L}_{G}^{HF}$ and $\mathcal{L}_{\text {align }}$, the coefficient of each loss is set to $1$.


\section{Experiments}
\label{sec:experiments}

\textbf{Datasets.}
We test the effectiveness of our method on low-shot datasets from various domains with different resolutions.
On 256 $\times$ 256 resolution, we use Animal Face Dogs and Cat~\cite{si2011learning}, as well as 100-shot-Panda, Obama, and Grumpy\_cat~\cite{DiffAug}.
On 512 $\times$ 512 resolution, we use Anime-Face, Art Paintings, Moongate, Flat-colored, and Fauvism-still-life~\cite{liu2021towards}.
On 1024 $\times$ 1024 resolution, we use Pokemon, Skulls, Shells, MetFace~\cite{karras2020training} and BrecaHAD~\cite{aksac2019brecahad}.
These datasets contain a limited number of samples (mostly less than 1,000) and cover art paintings, realistic photos, human faces, etc.
For datasets that are not strictly equal to the corresponding resolution, we resize them to the closest resolution in implementation.
Besides, we use AnimalFace HQ (AFHQ) datasets~\cite{choi2020stargan} to evaluate the performance of our model when training with more data ($\sim$5k).

\textbf{Evaluation metrics and baseline.}
We adopt two common metrics to evaluate the {synthesize} quality: Fr\'{e}chet Inception Distance (FID)~\cite{heusel2017gans} and Kernel Inception Distance (KID)~\cite{binkowski2018demystifying}.
The lower FID and KID is, the better the generation quality is.
FID quantifies the distance between the distribution of the generated and the real images.
KID, which is designed unbiased, has been proven more descriptive for small datasets~\cite{karras2020training}, note that all KID scores reported in our paper need to $\times 10^{-3}$ following~\cite{karras2020training}.
Following~\cite{liu2021towards}, we calculate FID and KID by measuring the distance between all available training images and 5k generated images.
We also provide the {LPIPS~\cite{zhang2018unreasonable}}, IS~\cite{salimans2016improved}, Precision, Recall~\cite{kynkaanniemi2019improved}, Density, Coverage~\cite{naeem2020reliable} results in the appendix.

We compare our model with:
1) the state-of-the-art generative model StyleGAN2~\cite{karras2020analyzing}, and SWAGAN~\cite{gal2021swagan}, which incorporates wavelet into StyleGAN2;
2) data augmentation-based approaches that is designed for training GANs with limited data, \emph{i.e.}, ADA~\cite{karras2020training}, DiffAug~\cite{DiffAug}, APA~\cite{jiang2021deceive};
3) the state-of-the-art few-shot generative model FastGAN~\cite{liu2021towards}.
We reimplement all baselines with their released official code under consistent settings for a fair comparison.
{Implementation} details of baseline models are given in the appendix.

\textbf{Implementation details.}
We choose the current state-of-the-art few-shot generative model FastGAN~\cite{liu2021towards} as the backbone and implement our proposed techniques upon it.
all other settings remain the same as ~\cite{liu2021towards}.
We decompose the intermediate 8 $\times$ 8, 16 $\times$ 16, 32 $\times$ 32 features of G and D into frequency components for our frequency skip connection, high-frequency discriminator, and high-frequency alignment.
More implement details are given in the appendix.

\begin{figure}
	\centering
	\vspace{-0.2cm}
	\includegraphics[width=\linewidth]{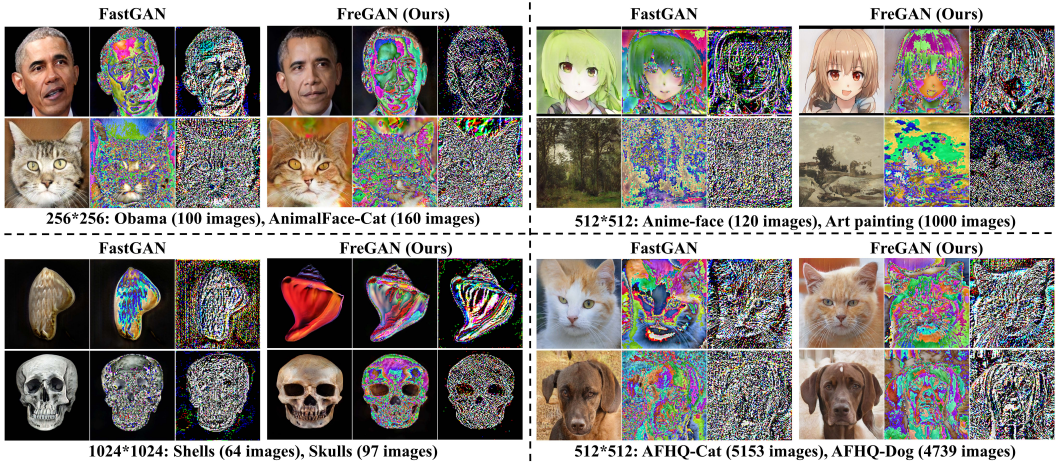}
	\vspace{-0.4cm}
	\caption{\textbf{Qualitative comparison results of our FreGAN and baseline FastGAN.} The images from left to right are generated images, low-frequency, and high-frequency components, respectively.
    Our FreGAN improves the overall quality of generated images and raises the model's frequency awareness, encouraging the generator to produce precise high-frequency signals with fine details.}
	\label{fig:visallmain}
	\vspace{-5mm}
\end{figure}

\subsection{Main Results}

\textbf{Quantitative comparison on datasets with limited data amounts.} The quantitative comparison results of our FreGAN and baseline methods on different resolutions are given in Tab.~\ref{tab:res256}, Tab.~\ref{tab:res512} and Tab.~\ref{tab:res1024}.
We save the best training snapshots of each method and generate 5k images to compute FID and KID.
The whole training set is adopted as the referenced distribution.
We can observe from the results that, although evaluated on various datasets that have different resolutions and data amounts, our proposed FreGAN achieves superior performance on all these datasets.
Our FreGAN consistently improves both FID and KID metrics on 14 of the 15 datasets, demonstrating the effectiveness and generalizability of our proposed techniques.
Notably, for those datasets with extremely limited data (less than 100), \ie, Flat (Tab.~\ref{tab:res512}), Shells and Skulls (Tab.~\ref{tab:res1024}),
our method improves the performance more significantly, \eg, the FID from 216.27 to \textbf{178.10} on Flat and from 101.94 to \textbf{86.12} on Skulls, and the corresponding KID is improved doubled, further reflecting our model's potential for training GANs with extremely limited data.
More quantitative results are presented in the appendix.
\begin{table}[tb!]
\centering
\small
\caption{The FID (lower is better) and KID (lower is better) scores of our method compared to state-of-the-art methods on \textbf{ \bm{$256 \times 256$} datasets with limited data amounts}.}
\vspace{-0.1cm}
\begin{tabularx}{\textwidth}{c*{10}{|Y}}
\Xhline{1pt}
                                        & \multicolumn{4}{c|}{{Animal Face}}                                            & \multicolumn{6}{c}{{100-shot}} \\ \cline{2-11}
                                        & \multicolumn{2}{c}{Dog (389 imgs)}   & \multicolumn{2}{|c}{Cat (160 imgs)}      & \multicolumn{2}{|c}{Panda}     & \multicolumn{2}{|c}{Obama}      & \multicolumn{2}{|c}{Grumpy\_cat}\\ \cline{2-11}
Method                                  & FID           & KID         & FID         & KID       & FID       & KID       & FID       & KID       & FID       & KID   \\ \Xhline{0.6pt}
StyleGAN2~\cite{karras2020analyzing}    & 113.86        & 91.31       & 79.04       & 34.43     & 18.05     & 7.40      & 69.01     & 52.63     & 35.00     & 11.01 \\
SWAGAN~\cite{gal2021swagan}             & {82.47}       & {80.46}     & {59.71}     & {19.35}   & {27.55}   & {11.92}   & {71.05}   & {55.82}   & {38.44}   & {17.03} \\
ADA~\cite{karras2020training}           & 55.48         & 18.42       & 37.95       & 6.43      & 14.17     & 6.53      & 43.17     & 13.23     & 43.80     & 45.01 \\
APA~\cite{jiang2021deceive}             & 81.16         & 26.42       & 42.60       & 7.97      & 19.21     & 10.80     & 42.97     & 15.71     & 28.10     & 5.53  \\
DiffAug~\cite{DiffAug}                  & 61.34         & 24.51       & 41.84       & 12.27     & 11.52     & 3.57      & 48.85     & 23.31     & 26.89     & 9.42  \\
FastGAN~\cite{liu2021towards}           & 52.46         & 18.22       & 33.85       & 4.99      & 9.70      & 1.60      & 35.80     & 5.50      & 25.75     & {\bf3.41}  \\    \Xhline{0.4pt}
FreGAN (Ours)                         & {\bf47.85}    & {\bf13.49}  & {\bf31.05}  & {\bf2.44} & {\bf8.97} & {\bf0.91} & {\bf33.39}& {\bf3.76} & {\bf24.93}  & 3.89\\ \Xhline{1pt}
\end{tabularx}
\label{tab:res256}
\vspace{-0.41cm}
\end{table}
\begin{table}[tb!]
\centering
\small
\caption{The FID (lower is better) and KID (lower is better) scores of our method compared to state-of-the-art methods on \textbf{ \bm{$512 \times 512$} datasets with limited data amounts}.}
\vspace{-0.1cm}
\begin{tabularx}{\textwidth}{c*{10}{|Y}}
\Xhline{1pt}

                                        & \multicolumn{2}{c}{AnimeFace}   & \multicolumn{2}{|c}{ArtPainting}      & \multicolumn{2}{|c}{Moongate}     & \multicolumn{2}{|c}{Flat}      & \multicolumn{2}{|c}{Fauvism}\\
                                        & \multicolumn{2}{c}{120 imgs}   & \multicolumn{2}{|c}{1000 imgs}      & \multicolumn{2}{|c}{136 imgs}     & \multicolumn{2}{|c}{36 imgs}      & \multicolumn{2}{|c}{124 imgs}\\ \cline{2-11}
Method                                  & FID       & KID       & FID       & KID       & FID         & KID       & FID         & KID        & FID       & KID    \\ \Xhline{0.6pt}
StyleGAN2~\cite{karras2020analyzing}    & 183.44    & 242.83    & 100.35    & 113.75    & 288.25      & 93.14     & 285.61      & 214.47     & 299.15    & 220.14 \\
SWAGAN~\cite{gal2021swagan}             & 189.71    & 216.39    & 56.95     & 22.50     & 302.72      & 99.47     & 293.94      & 232.53     & 291.66    & 226.21 \\
ADA~\cite{karras2020training}           & 59.67     & 16.02     & 46.38     & 12.26     & 149.06      & 43.21     & 248.46      & 62.89      & 201.99    & 86.64  \\
APA~\cite{jiang2021deceive}             & 58.38     & 15.73     & 47.23     & 10.60     & 193.67      & 50.52     & 233.52      & 166.53     & 197.47    & 66.13  \\
DiffAug~\cite{DiffAug}                  & 135.85    & 148.51    & 49.25     & 18.42     & 136.12      & 48.04     & 340.14      & 247.41     & 223.58    & 117.10 \\
FastGAN~\cite{liu2021towards}           & 55.87     & 11.17     & 45.06     & 10.26     & 114.79      & 23.57     & 216.27      & 36.88      & 178.42    & 58.01  \\ \Xhline{0.4pt}
FreGAN (Ours)                           & {\bf50.19}& {\bf4.58} & {\bf43.13}& {\bf9.71} & {\bf107.13}& {\bf15.58} & {\bf178.10} & {\bf18.35} &{\bf171.95}& {\bf49.81}  \\ \Xhline{1pt}
\end{tabularx}
\label{tab:res512}
\vspace{-0.41cm}
\end{table}
\begin{table}[tb!]
\centering
\small
\caption{The FID (lower is better) and KID (lower is better) scores of our method compared to state-of-the-art methods on \textbf{ \bm{$1024 \times 1024$} datasets with limited data amounts}.}
\vspace{-0.1cm}
\begin{tabularx}{\textwidth}{c*{10}{|Y}}
\Xhline{1pt}

                                        & \multicolumn{2}{c}{Shells}   & \multicolumn{2}{|c}{Skulls}      & \multicolumn{2}{|c}{Pokemon}     & \multicolumn{2}{|c}{BrecaHAD}      & \multicolumn{2}{|c}{MetFace}\\
                                        & \multicolumn{2}{c}{64 imgs}  & \multicolumn{2}{|c}{97 imgs}     & \multicolumn{2}{|c}{833 imgs}    & \multicolumn{2}{|c}{162 imgs}      & \multicolumn{2}{|c}{1336 imgs}\\ \cline{2-11}
Method                                  & FID        & KID          & FID       & KID       & FID       & KID       & FID       & KID       & FID       & KID   \\ \Xhline{0.6pt}
StyleGAN2~\cite{karras2020analyzing}    & 133.31     & 33.36        & 234.54    & 209.22    & 161.28    & 161.98    & 174.07    & 176.32    & 66.97     & 55.53 \\
SWAGAN~\cite{gal2021swagan}             & 185.96     & 85.25        & 203.49    & 178.96    & 80.94     & 68.02     & 162.53    & 119.64    & 31.56     & 13.96  \\
ADA~\cite{karras2020training}           & 133.22     & 29.12        & 97.05     & 12.33     & 66.41     & -         & 76.67     & 21.38     & 24.74     & 10.23 \\
APA~\cite{jiang2021deceive}             & 136.52     & 58.77        & 99.46     & 12.74     & 51.05     & 59.29     & 75.89     & 25.08     & 26.03     & {\bf5.58}  \\
DiffAug~\cite{DiffAug}                  & 151.94     & 54.73        & 124.23    & 38.12     & 62.73     & 50.68     & 93.71     & 31.62     & 27.45     & 11.55 \\
FastGAN~\cite{liu2021towards}           & 141.71     & 37.00        & 101.94    & 12.10     & 44.96     & 17.31     & 59.80     & 7.24      & 26.80     & 7.08  \\ \Xhline{0.4pt}
FreGAN (Ours)                         & {\bf125.77}& {\bf20.58}   & {\bf86.12}& {\bf5.47} & {\bf38.88}& {\bf10.42}& {\bf54.88}& {\bf3.41} & {\bf25.42}  & 5.93\\ \Xhline{1pt}
\end{tabularx}
\label{tab:res1024}
\vspace{-0.55cm}
\end{table}

\textbf{Qualitative Comparison.} The qualitative results of FastGAN and our FreGAN on various datasets are illustrated in Fig.~\ref{fig:visallmain}.
For each dataset in Fig.~\ref{fig:visallmain}, from left to right are generated images, the visualization of the low and high-frequency components of the generated images.
The images generated by FastGAN contain unsatisfactory artifacts and some of them are incongruous, \eg, the generated images of cat and dog in the bottom right of Fig.\ref{fig:visallmain}, the cat has artifacts around the
head, and the dog's ears are distorted.
Our FreGAN significantly facilitates image quality in coordination, rationality, and fine details.
As can been seen from Fig.~\ref{fig:visallmain}, the human face Obama generated by our FreGAN is more photorealistic, the details of the anime face, such as eye color and hair texture, are more realistic, and the
synthesized animal faces of cats and dogs are also more plausible.
Besides, the frequency components of the images generated by our FreGAN contain wealthier details.
For example, the generated image of AnimalFace-Cat has a richer background, and the generated image of Skulls has more clear contours of the eye and nose.
Such observation reflects that the proposed FreGAN: 1) ameliorates the quality of generated images under limited data;
2) raises frequency awareness of {synthesizing} high-frequency signals with richer fine details of images;
and 3) takes full advantage of limited data's frequency information.
More qualitative results are given in the appendix.

\textbf{Effectiveness under datasets with more data.}
To investigate the effectiveness of our FreGAN more comprehensively, we evaluate the performance on datasets with more training data,
\ie, AnimalFace-HQ (AFHQ)~\cite{choi2020stargan}, which includes 3 sub-datasets with close to 5k images, the results are shown in Tab.~\ref{tab:resMS}.
Similarly, our method yields compelling improvements on both FID and KID metrics when training with more data.
Combined with the generated images in Fig.~\ref{fig:visallmain}, the results further validate our FreGAN's contribution to the {synthesize} quality.
Our method boosts the performance under different amounts of data, suggesting the generalization of our model.
\begin{table}[tb!]
\vspace{-0.1cm}
\centering
\small
\caption{The FID (lower is better) and KID (lower is better) scores of our method compared to the state-of-the-art FastGAN on \textbf{AFHQ~\cite{choi2020stargan} datasets with more training data ($\sim$5k)}.}
\vspace{-0.12cm}
\begin{tabularx}{\textwidth}{c*{6}{|Y}}
\Xhline{1pt}
                                        & \multicolumn{2}{c}{AFHQ-Cat (5153 imgs)}   & \multicolumn{2}{|c}{AFHQ-Dog (4739 imgs)}      & \multicolumn{2}{|c}{AFHQ-Wild (4738 imgs)} \\ \cline{2-7}
Method                                  & FID           & KID                   & FID           & KID                       & FID           & KID       \\ \Xhline{0.6pt}
FastGAN~\cite{liu2021towards}           & 10.17         & 4.91                  & 25.36         & 14.29                     & 7.30          & 1.93      \\
+Ours                                   & {\bf6.62}     & {\bf1.95}             & {\bf20.75}    & {\bf11.45}                & {\bf6.37}     & {\bf1.31} \\ \Xhline{1pt}
\end{tabularx}
\label{tab:resMS}
\vspace{-0.38cm}
\end{table}

\subsection{Ablation Studies}
\label{sec:ablationstudy}
\textbf{Ablation studies on variants of FreGAN.} There are three ingredients of our FreGAN, \ie, the high-frequency discriminator (HFD), high-frequency alignments (HFA), and frequency skip connection (FSC).
We evaluate the efficacy of each component by removing each of them from the full version of our FreGAN.
We choose one from each of the different resolution datasets, \ie, 100-shot-Obama, Anime face and pokemon for 256, 512 and 1024 resolution, respectively.
As shown in Tab.~\ref{tab:ablationmain}, removing any of the three techniques leads to a performance drop, reflecting the contribution of each component.
Still, all these variants outperform baseline FastGAN on both FID and KID, which implies that the combination of different components of our method consistently boosts model performance.
Moreover, the performance drops the most when removing the HFD module, which is reasonable because the HFD raises the frequency awareness of G and D, and the frequency awareness of D serves as a self-supervision to guide G to {synthesize} adequate and reasonable frequency signals.
{Qualitative comparison results of ablation studies are given in the appendix.}
\begin{table}[tb!]
\centering
\small
\caption{\textbf{Ablation studies on different components of our FreGAN.} We remove each component to evaluate the efficacy of the three ingredients of our method, \ie, HFD, HFA, and FSC.
The ``Full'' represents the the full version combining all three techniques used in the main experiments.}
\vspace{-0.12cm}
\begin{tabularx}{\textwidth}{c*{6}{|Y}}
\Xhline{1pt}
                                        & \multicolumn{2}{c}{100-shot-Obama (256 $\times$ 256)}   & \multicolumn{2}{|c}{Anime Face (512 $\times$ 512)}      & \multicolumn{2}{|c}{Pokemon (1024 $\times$ 1024)}     \\ \cline{2-7}
Module                                  & FID       & KID                      & FID        & KID                     & FID                 & KID        \\ \Xhline{0.6pt}
Baseline                                & 35.80     & 5.50                     & 55.87      & 11.17                   & 44.96               & 17.31      \\ \Xhline{0.4pt}
w/o HFD                                 & 35.67     & 7.78                     & 55.17      & 8.16                    & 41.75               & 13.69      \\
w/o HFA                                 & 34.28     & 4.60                     & 54.40      & 10.70                   & 40.27               & 12.53      \\
w/o FSC                                 & 33.52     & 4.18                     & 51.15      & 4.83                    & 39.41               & 11.13      \\ \Xhline{0.4pt}
Full                                    & {\bf33.39}& {\bf3.76}                & {\bf50.19} & {\bf4.58}               & {\bf38.88}          & {\bf10.42} \\ \Xhline{1pt}
\end{tabularx}
\label{tab:ablationmain}
\vspace{-0.31cm}
\end{table}

\textbf{Ablation studies on different scale of features.} We employ our proposed HFD and HFA on multi-scale features of G and D, namely, 8, 16, and 32 scales of features.
Here we provide the ablation studies on different scales in Tab.~\ref{tab:ablationScale}.
It can be seen that performing HFD and HFA on multi-scale features boosts the model performance.
Besides, when only performing HFD and HFA on single-scale features, the obtained results still outperform the FastGAN baseline, suggesting the effectiveness of HFA and HFD.
Notably, despite adding more scales of features may bring further performance advancement, the required additional convolutional and downsampling layers increases for higher scales features(\eg, 128, 256), bringing non-negligible computational costs.

\textbf{Ablation studies on different frequency components.} Three high-frequency components are obtained from wavelet transformation on the features, \ie, $LH$, $HL$, and $HH$.
Each of them encodes different details of features as shown in Fig.~\ref{fig:frequency}, we sum them to fuse all the detail information for further operation in our main experiments.
Here we conduct experiments on the three components respectively to verify their contribution and the necessity of fusing them.
As shown in Tab.~\ref{tab:ablationScale}, each high-frequency component contributes to the model performance compared with the baseline, and fusing them can better promote the generation quality.
\begin{table}[tb!]
\vspace{-0.1cm}
\centering
\small
\caption{\textbf{Ablation studies on the different scales of HFD and HFA on AnimalFace-Dog dataset.} The numbers after ``Feat'' indicate the scale of features. ``LH'' and ``HL'' denote the high-frequency components.
The ``Full'' represents the full version of our FreGAN used in the main experiments.}
\vspace{-0.1cm}
\begin{tabularx}{\textwidth}{c*{7}{|Y}} \Xhline{1pt}
Module      &Metric     & Baseline      & Feat8         & Feat8 + 16             & LH               & LH + HL      & Full \\ \Xhline{0.6pt}
HFD         &FID        & 52.46         & 51.86         & 50.63                 & 52.15            & 51.79        & {\bf47.85}  \\
HFD         &KID        & 18.22         & 17.13         & 16.28                 & 17.94            & 16.57        & {\bf13.49}  \\ \Xhline{0.4pt}
HFA         &FID        & 52.46         & 52.23         & 51.93                 & 51.60            & 49.71        & {\bf47.85}  \\
HFA         &KID        & 18.22         & 17.15         & 16.87                 & 16.92            & 15.87        & {\bf13.49}  \\ \Xhline{1pt}
\end{tabularx}
\label{tab:ablationScale}
\vspace{-0.41cm}
\end{table}
\begin{table}[tb!]
\centering
\small
\caption{The FID (lower is better) and KID (lower is better) scores of our method \textbf{combined with model regularization and attention mechanism techniques on \bm{$256 \times 256$} datasets.}}
\vspace{-0.1cm}
\begin{tabularx}{\textwidth}{c*{10}{|Y}}
\Xhline{1pt}
                                        & \multicolumn{4}{c|}{{Animal Face}}                                            & \multicolumn{6}{c}{{100-shot}} \\ \cline{2-11}
& \multicolumn{2}{c}{Dog (389 imgs)}   & \multicolumn{2}{|c}{Cat (160 imgs)}      & \multicolumn{2}{|c}{Panda}     & \multicolumn{2}{|c}{Obama}      & \multicolumn{2}{|c}{Grumpy\_cat}\\ \cline{2-11}
Method                                  & FID           & KID         & FID         & KID       & FID        & KID       & FID       & KID       & FID       & KID      \\ \Xhline{0.6pt}
Lecam~\cite{tseng2021regularizing}      & 54.88         & -           & 34.18       & -         & 10.16      & -         & 33.16     & -         & 24.93     & -        \\
+ Ours                                  & {\bf48.29}    & {\bf14.16}  & {\bf31.77}  & {\bf2.22} & {\bf8.87}  & {\bf1.06} & {\bf32.69}& {\bf4.99} & {\bf24.39}& {\bf2.36}\\ \Xhline{0.4pt}
MoCA~\cite{li2021prototype}             & 54.04         & 19.25       & 38.04       & 8.40      & 11.24      & 4.00      & 42.26     & 17.03     & 25.59     & 4.20     \\
+ Ours                                  & {\bf50.96}    & {\bf16.06}  & {\bf35.47}  & {\bf4.92} & {\bf9.05}  & {\bf1.13} & {\bf34.13}& {\bf5.53} & {\bf24.78}& {\bf3.11}\\ \Xhline{0.6pt}
\end{tabularx}
\label{tab:Compatibility}
\vspace{-0.31cm}
\end{table}
\subsection{Analysis on Compatibility and GAN Equilibrium}
\textbf{Compatibility of Our Model.}
Lecam~\cite{tseng2021regularizing} and MoCA~\cite{li2021prototype} exploit regularization and attention mechanism for training GANs under limited data, respectively.
We implement our proposed techniques on them to test the compatibility of our method.
We keep the original setting unchanged and the set the regularization weight to 0.1.
The FID results are given in Tab.~\ref{tab:Compatibility}, from which we can see that FreGAN can further boost the performance of MoCA and Lecam, demonstrating that our method is complementary
to the model regularization and attention mechanism methods.

\begin{figure}
	\centering
	\vspace{-0.1cm}
	\includegraphics[width=\linewidth]{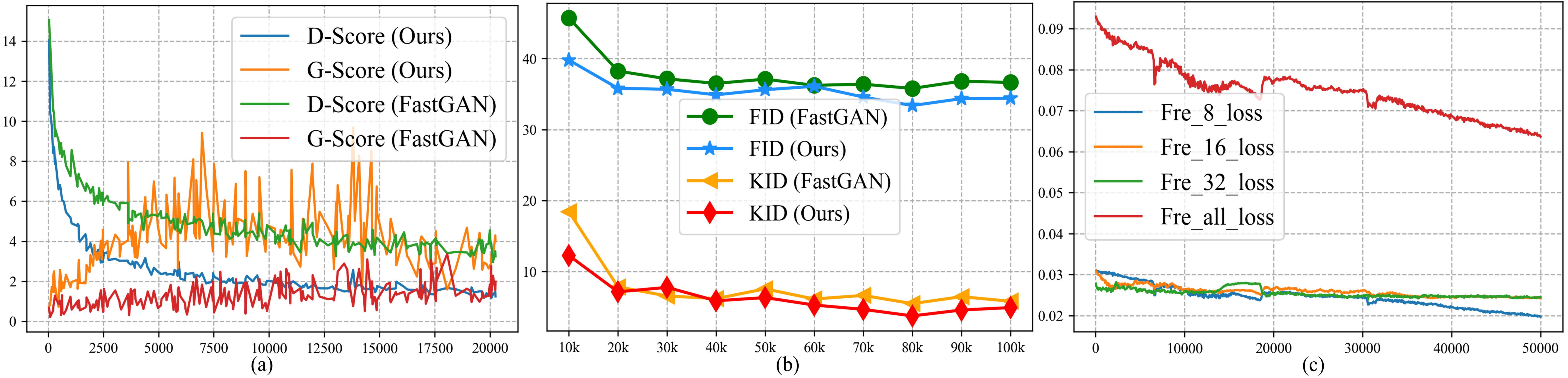}
	\vspace{-0.6cm}
	\caption{\textbf{Equilibrium is improved with our proposed techniques.} (a) Our generator can better deceive the discriminator. (b) Our FreGAN yields better performance throughout the training process.
    (c) The aligned HFA loss mitigates the asymmetrical information and facilitates the GAN equilibrium.}
	\label{fig:equlibrium}
	\vspace{-6mm}
\end{figure}

\textbf{GAN Equilibrium is improved.}
Our HFA module aligns the frequency components of real and generated images, guiding G to {synthesize} precise instead of arbitrary high-frequency signals.
Meanwhile, as a byproduct, the HFA mitigates the domain gap between G and D, alleviating the unhealthy competition.
As shown in Fig.~\ref{fig:equlibrium} (a), our discriminator converges to a better point, and our generator can better fool the discriminator, while the discriminator of FastGAN surpass the generator, thus providing less informative guidelines and degrading the {synthesize} quality.
Besides, we plot the FID and KID curves throughout the training process in Fig.~\ref{fig:equlibrium} (b), from which we can observe that our FreGAN are consistently better.
Moreover, we plot the multi-scale HFA loss curves in Fig.~\ref{fig:equlibrium} (c), where each line denotes the loss of each scale.
These curves indicate that the frequency signals are well aligned, lessening the domain gaps and promoting the GAN equilibrium.

\vspace{-4mm}
\section{Discussion}
\label{sec:discussion}

\textbf{Conclusions.}
In this paper, we propose a frequency-aware method for training GANs under limited data, \ie, FreGAN.
The proposed FreGAN ameliorates the {synthesize} quality by raising the model's frequency awareness, encouraging the model to pay more attention to frequency signals, especially high-frequency signals, which encode fine details of images.
We conduct extensive experiments on various datasets with different amounts of data and different resolutions to demonstrate the efficacy of our proposed method.
Qualitative results suggest that our model successfully makes the generator to generate precise high-frequency signals, facilitating high-quality image generation.
Quantitative results indicate that our method 1) substantially boosts the performance, especially when data is extremely limited (less than 100), and 2) is complementary to existing regularization and attention models.
Moreover, the proposed model alleviates the disequilibrium of GANs by lessening the frequency information gap.
In the future, we plan to implement our techniques on more backbones, \eg, StyleGAN2~\cite{karras2020analyzing} and apply our method to more applications.

\textbf{Limitations.}
Despite achieving significant improvements on various low-data datasets, our FreGAN still struggles in generating photorealistic images when given datasets with limited data but various contents, \eg, only dozens of images, and their contents vary widely. 
When the low-data datasets are imbalanced~\cite{8968753} or even long-tailed, the proposed method may fail to generalize, which is limited by the intrinsic reasons of the data distribution.
Developing more effective ways to train generative models with insufficient training data still requires more efforts.

\section*{Acknowledgment}
This work is supported by Shanghai Science and Technology Program ``Distributed and generative few-shot algorithm and theory research'' under Grant No. 20511100600 and ``Federated based cross-domain and cross-task incremental learning'' under Grant No. 21511100800, Natural Science Foundation of China under Grant No. 62076094, Chinese Defense Program of Science and Technology under Grant No.2021-JCJQ-JJ-0041, China Aerospace Science and Technology Corporation Industry-University-Research Cooperation Foundation of the Eighth Research Institute under Grant No.SAST2021-007.

{\small
\bibliographystyle{plain}
\bibliography{sections/egbib}
}
\newpage
\section*{Checklist}


\begin{enumerate}

\item For all authors...
\begin{enumerate}
  \item Do the main claims made in the abstract and introduction accurately reflect the paper's contributions and scope?
    \answerYes{}
  \item Did you describe the limitations of your work?
    \answerYes{} Section~\ref{sec:discussion}.
  \item Did you discuss any potential negative societal impacts of your work?
    \answerYes{} Section~\ref{sec:discussion}.
  \item Have you read the ethics review guidelines and ensured that your paper conforms to them?
    \answerYes{}
\end{enumerate}

\item If you are including theoretical results...
\begin{enumerate}
  \item Did you state the full set of assumptions of all theoretical results?
    \answerNA{}
        \item Did you include complete proofs of all theoretical results?
    \answerNA{}
\end{enumerate}

\item If you ran experiments...
\begin{enumerate}
  \item Did you include the code, data, and instructions needed to reproduce the main experimental results (either in the supplemental material or as a URL)?
    \answerYes{} Code is provided in the supplemental material. We will also make our code and pre-trained models publicly available \footnote{Our codes are available at \url{https://github.com/kobeshegu/FreGAN_NeurIPS2022}.}.
  \item Did you specify all the training details (e.g., data splits, hyperparameters, how they were chosen)?
    \answerYes{} Implement details in Section~\ref{sec:experiments}.
        \item Did you report error bars (e.g., with respect to the random seed after running experiments multiple times)?
    \answerNo{}
        \item Did you include the total amount of compute and the type of resources used (e.g., type of GPUs, internal cluster, or cloud provider)?
    \answerYes{} Implement details in Section~\ref{sec:experiments}.
\end{enumerate}

\item If you are using existing assets (e.g., code, data, models) or curating/releasing new assets...
\begin{enumerate}
  \item If your work uses existing assets, did you cite the creators?
    \answerYes{} Section~\ref{sec:experiments}.
  \item Did you mention the license of the assets?
    \answerNo{}
  \item Did you include any new assets either in the supplemental material or as a URL?
    \answerNo{}
  \item Did you discuss whether and how consent was obtained from people whose data you're using/curating?
    \answerYes{} Datasets in Section~\ref{sec:experiments}.
  \item Did you discuss whether the data you are using/curating contains personally identifiable information or offensive content?
    \answerNo{} The data we use contains no personally identifiable information or offensive content, the description of datasets is given in Section~\ref{sec:experiments}.
\end{enumerate}

\item If you used crowdsourcing or conducted research with human subjects...
\begin{enumerate}
  \item Did you include the full text of instructions given to participants and screenshots, if applicable?
    \answerNA{}
  \item Did you describe any potential participant risks, with links to Institutional Review Board (IRB) approvals, if applicable?
    \answerNA{}
  \item Did you include the estimated hourly wage paid to participants and the total amount spent on participant compensation?
    \answerNA{}
\end{enumerate}

\end{enumerate}

\newpage
\appendix


\section{Appendix}
\label{sec:appendix}
In this appendix, we first provide the broader impact of our method. Then we provide more implement details (Sec.~\ref{sec:implementdetails}), quantitative results of other metrics (Sec.~\ref{sec:quantitativemoredata} and Sec.~\ref{sec:quantitativeapp}), and qualitative results of generated images and {ablation studies} (Sec.~\ref{sec:qualitativeapp}) that is not elaborated in the main paper.
{We also provide the quantitative comparison results of our FreGAN and FastGAN on large-scale datasets in Sec.ref{sec:largedatasets}
Besides, we provide the latent space interpolation and nearest neighbors found from training images in Sec.~\ref{sec:diversityapp} for a more comprehensive analysis on the diversity of our generated images.
Lastly, more quantitative and qualitative comparison results of spectral properties are given in Sec.~\ref{sec:spectralanaapp}, demonstrating that our FreGAN is frequency-aware and can indeed produce realistic frequency signals.}

\textbf{Broader impact.}
{The proposed method enables data-efficient GANs training for high-quality image synthesize with limited data.
It benefits the practical implementation if GANs in various applications that without sufficient training samples, \emph{e.g.}, medical images and art paintings.
And our analysis of the effectiveness of frequency components in image synthesis may also extend the breadth and potential of approaches for training effective GANs from the frequency domain perspective.
Besides, being capable of generating plausible and photorealistic images, our method bring potential issue of image abuse and fraud with the generated fake images.
However, we believe that the rational use of such advanced technology can bring benefits to more fields like films and art production.}

\subsection{More implement details.}
\label{sec:implementdetails}
\paragraph{Implement details of our FreGAN.}
We perform Haar wavelet transformation on the intermediate 8 $\times$ 8, 16 $\times$ 16, 32 $\times$ 32 features of G and D.
The PyTorch-like pseudocode of Haar wavelet transformation is given in Algorithm.~\ref{alg:code}.
We perform FSC by wavelet unpooling the decomposed high-frequency components and feeding the reconstructed features to the subsequent layers.
For HFD, we aggregate the high-frequency components by adding $LH, HL, HH$ and then employ additional downsampling and convolutional layers to compute the output scores.
Specifically, the architecture resembles the original discriminator. The added layers on the high-frequency components include 2d convolutional, 2d Batch Normalization, and LeakyReLU layers.
The difference is that the HFD discriminates the input images' frequency information, raising the discriminator's frequency awareness.
For HFA, we align the summed frequency signals of real and generated intermediate features by minimizing Eq.4 in the main paper.
Notably, only 1-2 additional layers are added for each high-frequency component without requiring much computational cost.
We train our model for 100k iterations and save the checkpoints every 10k iterations.
The saved checkpoints are used to generate images for evaluation.
All experiments are run on 2 Tesla V100 GPUs with PyTorch framework, and our code will be made available online.
\begin{algorithm}[t]
\caption{Haar wavelet transformation pseudocode, PyTorch-like}
\label{alg:code}
\definecolor{codeblue}{rgb}{0.25,0.5,0.5}
\definecolor{codekw}{rgb}{0.85, 0.18, 0.50}
\lstset{
  backgroundcolor=\color{white},
  basicstyle=\fontsize{7.5pt}{7.5pt}\ttfamily\selectfont,
  columns=fullflexible,
  breaklines=true,
  captionpos=b,
  commentstyle=\fontsize{7.5pt}{7.5pt}\color{codeblue},
  keywordstyle=\fontsize{7.5pt}{7.5pt}\color{codekw},
}
\begin{lstlisting}[language=python]
def get_wav(in_channels, pool=True):
    # wavelet decomposition using conv2d
    harr_wav_L = 1 / np.sqrt(2) * np.ones((1, 2))
    harr_wav_H = 1 / np.sqrt(2) * np.ones((1, 2))
    harr_wav_H[0, 0] = -1 * harr_wav_H[0, 0]

    harr_wav_LL = np.transpose(harr_wav_L) * harr_wav_L
    harr_wav_LH = np.transpose(harr_wav_L) * harr_wav_H
    harr_wav_HL = np.transpose(harr_wav_H) * harr_wav_L
    harr_wav_HH = np.transpose(harr_wav_H) * harr_wav_H

    filter_LL = torch.from_numpy(harr_wav_LL).unsqueeze(0)
    filter_LH = torch.from_numpy(harr_wav_LH).unsqueeze(0)
    filter_HL = torch.from_numpy(harr_wav_HL).unsqueeze(0)
    filter_HH = torch.from_numpy(harr_wav_HH).unsqueeze(0)

    if pool:
        net = nn.Conv2d
    else:
        net = nn.ConvTranspose2d
    LL = net(in_channels, in_channels*2,
             kernel_size=2, stride=2, padding=0, bias=False,
             groups=in_channels)
    LH = net(in_channels, in_channels*2,
             kernel_size=2, stride=2, padding=0, bias=False,
             groups=in_channels)
    HL = net(in_channels, in_channels*2,
             kernel_size=2, stride=2, padding=0, bias=False,
             groups=in_channels)
    HH = net(in_channels, in_channels*2,
             kernel_size=2, stride=2, padding=0, bias=False,
             groups=in_channels)

    LL.weight.requires_grad = False
    LH.weight.requires_grad = False
    HL.weight.requires_grad = False
    HH.weight.requires_grad = False

    LL.weight.data = filter_LL.float().unsqueeze(0).expand(in_channels*2, -1, -1, -1)
    LH.weight.data = filter_LH.float().unsqueeze(0).expand(in_channels*2, -1, -1, -1)
    HL.weight.data = filter_HL.float().unsqueeze(0).expand(in_channels*2, -1, -1, -1)
    HH.weight.data = filter_HH.float().unsqueeze(0).expand(in_channels*2, -1, -1, -1)

    return LL, LH, HL, HH

class WavePool(nn.Module):
    def __init__(self, in_channels):
        super(WavePool, self).__init__()
        self.LL, self.LH, self.HL, self.HH = get_wav(in_channels)

    def forward(self, x):
        return self.LL(x), self.LH(x), self.HL(x), self.HH(x)

class WaveUnpool(nn.Module):
    def __init__(self, in_channels, option_unpool='cat5'):
        super(WaveUnpool, self).__init__()
        self.in_channels = in_channels
        self.option_unpool = option_unpool
        self.LL, self.LH, self.HL, self.HH = get_wav(self.in_channels, pool=False)

    def forward(self, LL, LH, HL, HH, original=None):
        if self.option_unpool == 'sum':
            return self.LL(LL) + self.LH(LH) + self.HL(HL) + self.HH(HH)
        elif self.option_unpool == 'cat5' and original is not None:
            return torch.cat([self.LL(LL), self.LH(LH), self.HL(HL), self.HH(HH), original], dim=1)
        else:
            raise NotImplementedError
\end{lstlisting}
\end{algorithm}

\paragraph{Implement details of baseline methods.}
We reimplement all the baseline methods with their official code for fair comparisons.
For StyleGAN2~\cite{karras2020analyzing}\footnote{\url{https://github.com/NVlabs/stylegan2}}, ADA~\cite{karras2020training}\footnote{\url{https://github.com/NVlabs/stylegan2-ada}}, DiffAug~\cite{DiffAug}\footnote{\url{https://github.com/mit-han-lab/data-efficient-gans}}, we keep most of the details unchanged, including style mixing regularization, path length regularization, exponential moving average of weights, non-saturating logistic loss with $R_1$ regularization. We show the discriminator for 2,000 kimg and use the best training snapshots of each model to generate 5k images for evaluation.
For FastGAN~\cite{liu2021towards}\footnote{\url{https://github.com/odegeasslbc/FastGAN-pytorch}}, similarly, we keep all the details unchanged and implement our proposed techniques upon it, we train both our FreGAN and FastGAN for 100k iterations, and we use the saved checkpoints to generate 5k images for evaluation.
All of our experiments are run on 2 Tesla V100 GPUs, using PyTorch 1.8.0, and CUDA 11.1.
When combining our proposed techniques upon Lecam~\cite{tseng2021regularizing} and MoCA~\cite{li2021prototype}, we consistently keep all the details unchanged and add our proposed techniques to them.
The coefficient of the regularization term is set as 0.1 following the original paper.
Notably, we use the official code and recommended parameter of MoCA.
However, we achieve 42.26 on the 100-shot-Obama dataset instead of 37.19 reported in the original paper.
We infer this is caused by randomness and hardware differences.
Nonetheless, our proposed method contributes to the performance and complements the attention mechanism-based method.

\paragraph{Datasets description.}
\begin{itemize}
  \item AFHQ dataset\footnote{\url{https://github.com/clovaai/stargan-v2}} dataset contain $\sim$5k training images of animal faces with 512 $\times$ resolution. The dataset is made available under the Creative Commons BY-NC 4.0 license.
  \item 100-shot datasets\footnote{\url{https://data-efficient-gans.mit.edu/datasets/}} contain various contents of images, and all the datasets contain 100 training images.
  They are ideal for verifying the quality of the generation in low-shot scenarios.
  \item MetFace\footnote{\url{https://github.com/NVlabs/metfaces-dataset}} dataset contains 1336 high-quality PNG images at 1024 $\times$ 1024 resolution. The dataset is made available under the Creative Commons BY-NC 2.0 license.
  \item BrecaHAD\footnote{\url{https://figshare.com/articles/dataset/BreCaHAD_A_Dataset_for_Breast_Cancer_Histopathological_Annotation_and_Diagnosis/7379186}} dataset contains 162 images for breast cancer histopathological annotation and diagnosis. Its texture and content are complex, thus is suited for evaluating GANs' performance under limited data, facilitating the exploration of data-efficient GANs for downstream tasks of the medical field.
  \item anime-face, art-paintings, moongate, flat, fauvism, shells, skulls\footnote{\url{https://drive.google.com/file/d/1aAJCZbXNHyraJ6Mi13dSbe7pTyfPXha0/view/}}. These datasets include 60 ~ 1000 images with different resolutions. Thus we adopt them for evaluating our model under limited data. We resize them to the closest resolution in implementation.
\end{itemize}
All of the datasets we used in this paper are open-sourced, and we use them only for academic research without any commercial purposes.

\subsection{More Quantitative comparison on datasets with limited data amounts.}
\label{sec:quantitativemoredata}
We evaluate the performance of our FreGAN and baseline models on more datasets with limited data amounts in Tab.~\ref{tab:res256FIDAppendix}, namely, Medici, Temple, Bridge, and Wuzhen, all of which contain only 100 training images. The resolution of these datasets is 256 $ \times $ 256.
The FID and KID results are consistent with the results in the main paper.
Our FreGAN achieves better performance compared with the baseline models, suggesting the effectiveness and generalization of our model.
\begin{table}[tb!]
\centering
\small
\caption{The FID (lower is better) and KID (lower is better) scores of our method compared to state-of-the-art methods on \textbf{ \bm{$256 \times 256$} datasets with limited data amounts}.}
\begin{tabularx}{\textwidth}{c*{8}{|Y}}
\Xhline{1pt}
 & \multicolumn{2}{c}{Medici (100 imgs)}& \multicolumn{2}{|c}{Temple (100 imgs)} &\multicolumn{2}{|c}{Bridge (100 imgs)} & \multicolumn{2}{|c}{Wuzhen (100 imgs)}  \\ \cline{2-9}
Method                                  & FID           & KID         & FID         & KID       & FID       & KID       & FID       & KID       \\ \Xhline{0.6pt}
StyleGAN2~\cite{karras2020analyzing}    & 66.36         & 41.16       & 73.35       & 46.76     & 116.40    & 71.51     & 135.40    & 116.46    \\
ADA~\cite{karras2020training}           & 44.21         & -           & 49.72       & -         & 72.07     & -         & 92.81     & 47.01     \\
APA~\cite{jiang2021deceive}             & 76.11         & -           & 41.38       & 10.33     & 189.74    & 98.45     & 102.10    & 45.92     \\
DiffAug~\cite{DiffAug}                  & 42.63         & 21.23       & 50.73       & 11.19     & 49.97     & 11.92     & 122.44    & 78.08     \\
FastGAN~\cite{liu2021towards}           & 38.47         & 12.63       & 36.01       & 4.95      & 46.82     & 8.25      & 67.98     & 15.13     \\ \Xhline{0.4pt}
FreGAN (Ours)                           & {\bf27.30}    & {\bf3.34}   & {\bf33.38}  & {\bf3.23} & {\bf44.18}& {\bf7.25} & {\bf59.89}& {\bf6.62} \\ \Xhline{1pt}
\end{tabularx}
\label{tab:res256FIDAppendix}
\end{table}

\begin{figure}
	\centering
	\vspace{-0.1cm}
	\includegraphics[width=\linewidth]{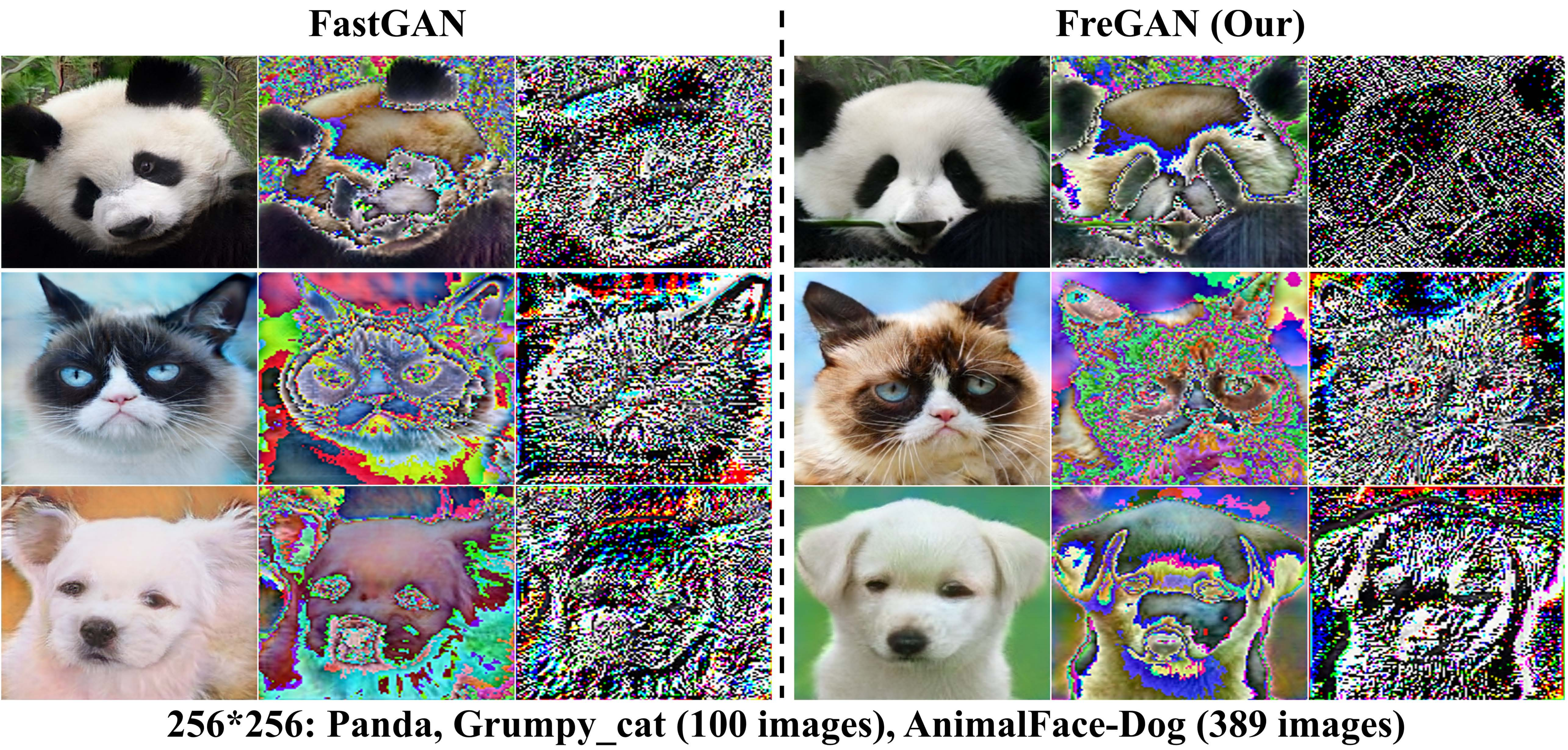}
	\vspace{-0.4cm}
	\caption{\textbf{Qualitative comparison results of our FreGAN and baseline FastGAN.} The images from left to right are generated images, low-frequency, and high-frequency components, respectively.
    Our FreGAN improves the overall quality of generated images and raises the model's frequency awareness, encouraging the generator to produce precise high-frequency signals with fine details.}
	\label{fig:Vis256Appendix}
	\vspace{-3mm}
\end{figure}

\begin{figure}
	\centering
	\vspace{-0.1cm}
	\includegraphics[width=\linewidth]{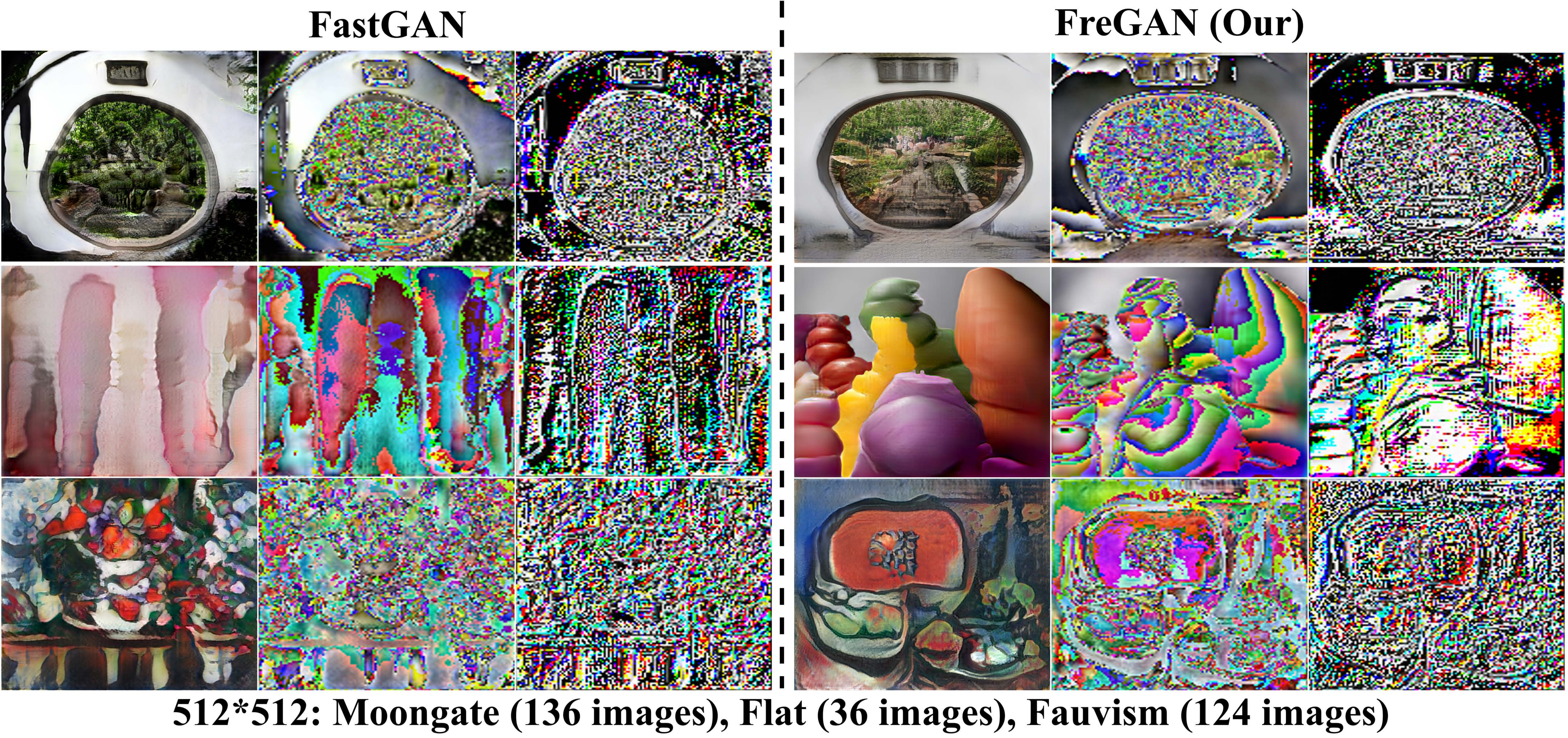}
	\vspace{-0.4cm}
	\caption{\textbf{Qualitative comparison results of our FreGAN and baseline FastGAN.} The images from left to right are generated images, low-frequency, and high-frequency components, respectively.
    Our FreGAN improves the overall quality of generated images and raises the model's frequency awareness, encouraging the generator to produce precise high-frequency signals with fine details.}
	\label{fig:Vis512Appendix}
	\vspace{-3mm}
\end{figure}

\begin{figure}
	\centering
	\vspace{-0.1cm}
	\includegraphics[width=\linewidth]{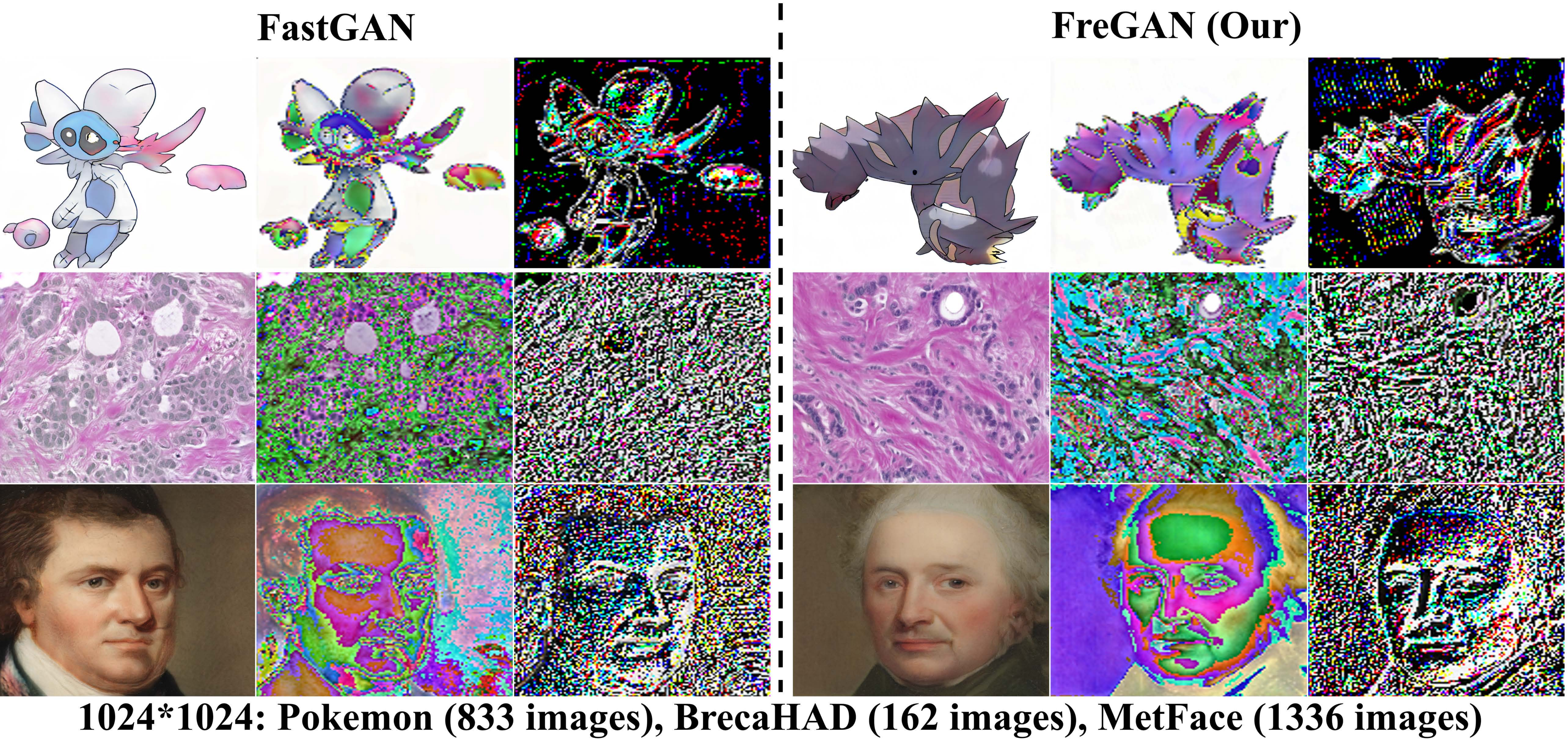}
	\vspace{-0.4cm}
	\caption{\textbf{Qualitative comparison results of our FreGAN and baseline FastGAN.} The images from left to right are generated images, low-frequency, and high-frequency components, respectively.
    Our FreGAN improves the overall quality of generated images and raises the model's frequency awareness, encouraging the generator to produce precise high-frequency signals with fine details.}
	\label{fig:Vis1024Appendix}
	\vspace{-3mm}
\end{figure}

\subsection{More Quantitative Comparison on Datasets with limited data amounts.}
\label{sec:quantitativeapp}
In addition to the two common-used metrics, \ie, FID and KID, we also compute the  Precision, Recall~\cite{kynkaanniemi2019improved}, Density, Coverage~\cite{naeem2020reliable}, Inception Score (IS)~\cite{salimans2016improved}, and {LPIPS~\cite{zhang2018unreasonable}} results in Tab.~\ref{tab:res256PrecisionAppendix1}, Tab.~\ref{tab:res256PrecisionAppendix2}, Tab.~\ref{tab:Precisionres512Appendix}, Tab.~\ref{tab:Precisionres1024Appendix}, Tab.~\ref{tab:PrecisionresMSAppendix},
Tab.~\ref{tab:Diversity256Appendix1}, Tab.~\ref{tab:res256DiversityAppendix2}, Tab.~\ref{tab:res512DiversityAppendix}, Tab.~\ref{tab:res1024DiversityAppendix}, Tab.~\ref{tab:DensityresMSAppendix},
Tab.~\ref{tab:res256ISAppendix1}, Tab.~\ref{tab:res256ISAppendix2}, Tab.~\ref{tab:res512ISAppendix}, Tab.~\ref{tab:res1024ISAppendix}, Tab.~\ref{tab:ISresMSAppendix}, Tab.~\ref{tab:res256LPIPS1}, Tab.~\ref{tab:res256LPIPS2}, Tab.~\ref{tab:res512LPIPS}, Tab.~\ref{tab:res1024LPIPS}.

Precision is quantified by calculating if the generated images are within the estimated manifold of real images, and symmetrically, Recall is quantified by calculating if the real images are within the estimated manifold of generated images~\cite{kynkaanniemi2019improved}.
Precision evaluates the probability of generated images falling into the real distribution, and Recall is the opposite.
Density and Coverage evaluate the fidelity and diversity of generative models, respectively~\cite{naeem2020reliable}, they are claimed successfully detect two identical distributions and are not robust against outliers.
Specifically, we use the generated 5k images of each model corresponding to the best FID results.
The whole training images are used as the referenced distribution.
We set the nearest k as 5 and compute the Precision, Recall, Density, and Coverage based on the official code of~\cite{naeem2020reliable}\footnote{\url{https://github.com/clovaai/generative-evaluation-prdc}}.
We split the generated into 10 parts for IS and report the mean and standard deviation of the calculated IS scores.
{We compute LPIPS~\cite{zhang2018unreasonable}\footnote{\url{https://github.com/richzhang/PerceptualSimilarity}} of all the paired of 5k images generated by each methods and report the average and standard deviation of LPIPS scores~\cite{sushko2021generating}.}

We can observe from these tables that:
1) KID and FID can consistently reflect the {synthesize} quality under limited data;
2) The Precision and Recall serve as supplementary metrics of evaluating generative GANs. They can reflect the distance between the generated and real distributions.
However, they may be biased when training data is limited since the model tends to overfit when given limited data, the model simply replicates the training images and achieves high Precision and Recall;
3) Evaluating the fidelity of generated images, the Density metric is more consistent with FID and KID, while the Coverage is not suited for evaluating the diversity as the {synthesized} 5k images cover the original training images, leading high value of the Coverage, \eg, equals to 1 on many datasets and baseline models;
4) Despite may be biased, our proposed FreGAN achieves advanced performance on most of the used datasets and the adopted evaluation metrics, demonstrating the effectiveness and superiority of our method;
5) {Notably, the recall is very low for all datasets and methods. We infer that this is possibly due to the low diversity of the training and the fact that recall is not suitable for evaluation in scenarios with limited data};
6) Devising indicative evaluation metrics for few-shot image generation tasks remains an open and tricky problem because one has to consider the degree of overfitting, the fidelity, and the diversity.

\begin{table}[tb!]
\centering
\small
\caption{The Precision (P) (higher is better) and Recall (R) (higher is better) scores of our method compared to state-of-the-art methods on \textbf{ \bm{$256 \times 256$} datasets with limited data amounts}.}
\begin{tabularx}{\textwidth}{c*{10}{|Y}}
\Xhline{1pt}
                                        & \multicolumn{4}{c|}{{Animal Face}}                                            & \multicolumn{6}{c}{{100-shot}} \\ \cline{2-11}
                                        & \multicolumn{2}{c}{Dog (389 imgs)}   & \multicolumn{2}{|c}{Cat (160 imgs)}      & \multicolumn{2}{|c}{Panda}     & \multicolumn{2}{|c}{Obama}      & \multicolumn{2}{|c}{Grumpy\_cat}\\ \cline{2-11}
Method                                  & P            & R          & P          & R         & P         & R         & P         & R         & P         & R        \\ \Xhline{0.6pt}
StyleGAN2~\cite{karras2020analyzing}    & 0.33         & 0.06       & 0.40       & 0.04      & 0.61      & 0.06      & 0.47     & {\bf0.16}  & 0.74     & 0.10      \\
ADA~\cite{karras2020training}           & 0.78         & {\bf0.50}  & 0.78       & 0.13      & 0.61      & 0.06      & 0.90     & 0.00       & 0.72     & 0.00      \\
APA~\cite{jiang2021deceive}             & 0.75         & 0.08       & 0.78       & 0.13      & 0.48      & 0.14      & 0.89     & 0.00       & 0.88     & 0.05      \\
DiffAug~\cite{DiffAug}                  & 0.79         & 0.31       & 0.87       & 1.05      & 0.79      & 0.11      & 0.85     & 0.00       & {\bf0.97}& 0.01      \\
FastGAN~\cite{liu2021towards}           & {\bf0.88}    & {\bf0.50}  & 0.87       & {\bf0.25} & {\bf0.91} & 0.10      & 0.94     & 0.10       & 0.93     & {\bf0.13} \\    \Xhline{0.4pt}
FreGAN (Ours)                           & 0.86         & 0.38       & {\bf0.90}  & 0.24      & {\bf0.91} & {\bf0.15} & {\bf0.96}& 0.11       & 0.94     & 0.03      \\ \Xhline{1pt}
\end{tabularx}
\label{tab:res256PrecisionAppendix1}
\end{table}

\begin{table}[tb!]
\centering
\small
\caption{The Precision (P) (higher is better) and Recall (R) (higher is better) scores of our method compared to state-of-the-art methods on \textbf{ \bm{$256 \times 256$} datasets with limited data amounts}.}
\begin{tabularx}{\textwidth}{c*{8}{|Y}}
\Xhline{1pt}
                                        & \multicolumn{2}{c}{Medici (100 imgs)}& \multicolumn{2}{|c}{Temple (100 imgs)} &\multicolumn{2}{|c}{Bridge (100 imgs)} & \multicolumn{2}{|c}{Wuzhen (100 imgs)}  \\ \cline{2-9}
Method                                  & P            & R           & P          & R         & P         & R         & P         & R        \\ \Xhline{0.6pt}
StyleGAN2~\cite{karras2020analyzing}    & 0.28         & 0.02        & 0.47       & {\bf0.06} & 0.79      & 0.00      & 0.09      & 0.04     \\
ADA~\cite{karras2020training}           & -            & -           & -          & -         & -         & -         & 0.50      & 0.00     \\
APA~\cite{jiang2021deceive}             & -            & -           & 0.91       & 0.01      & 0.25      & 0.00      & 0.44      & 0.06     \\
DiffAug~\cite{DiffAug}                  & 0.60         & 0.00        & 0.88       & {\bf0.06} & 0.89      & {\bf0.14} & 0.30      & 0.01     \\
FastGAN~\cite{liu2021towards}           & 0.87         & 0.00        & 0.91       & 0.05      & 0.88      & 0.09      & 0.80      & {\bf0.15}\\ \Xhline{0.4pt}
FreGAN (Ours)                           & {\bf0.93}    & {\bf0.03}   & {\bf0.93}  & 0.02      & {\bf0.90} & 0.04      & {\bf0.85} & 0.09     \\ \Xhline{1pt}
\end{tabularx}
\label{tab:res256PrecisionAppendix2}
\end{table}

\begin{table}[tb!]
\centering
\small
\caption{The Precision (P) (higher is better) and Recall (R) (higher is better) scores of our method compared to state-of-the-art methods on \textbf{ \bm{$512 \times 512$} datasets with limited data amounts}.}
\begin{tabularx}{\textwidth}{c*{10}{|Y}}
\Xhline{1pt}
                            & \multicolumn{2}{c}{AnimeFace}   & \multicolumn{2}{|c}{ArtPainting}      & \multicolumn{2}{|c}{Moongate}     & \multicolumn{2}{|c}{Flat}      & \multicolumn{2}{|c}{Fauvism}\\
                            & \multicolumn{2}{c}{120 imgs}   & \multicolumn{2}{|c}{1000 imgs}      & \multicolumn{2}{|c}{136 imgs}     & \multicolumn{2}{|c}{36 imgs}     & \multicolumn{2}{|c}{124 imgs}\\ \cline{2-11}
Method                                  & P             & R          & P          & R         & P        & R         & P        & R        & P        & R        \\ \Xhline{0.6pt}
StyleGAN2~\cite{karras2020analyzing}    & 0.86          & 0.00       & 0.34       & 0.01      & 0.63     & {\bf0.16} & 0.62     & 0.00     & 0.35     & 0.00     \\
ADA~\cite{karras2020training}           & 0.89          & 0.03       & 0.71       & 0.35      & 0.54     & 0.01      & 0.75     & 0.00     & 0.76     & 0.00     \\
APA~\cite{jiang2021deceive}             & 0.91          & 0.03       & 0.67       & {\bf0.38} & 0.46     & 0.00      & 0.50     & 0.00     & 0.73     & 0.00     \\
DiffAug~\cite{DiffAug}                  & 0.58          & 0.00       & 0.74       & 0.20      & 0.70     & 0.00      & 0.86     & 0.00     & 0.48     & 0.00     \\
FastGAN~\cite{liu2021towards}           & 0.89          & 0.12       & 0.81       & 0.32      & {\bf0.71}& 0.02      & 0.74     & {\bf0.03}& {\bf0.84}& {\bf0.02}\\    \Xhline{0.4pt}
FreGAN (Ours)                           & {\bf0.93}     & {\bf0.13}  & {\bf0.83}  & 0.33      & {\bf0.71}& 0.06      & {\bf0.90}& {\bf0.03}& 0.82     & {\bf0.02}\\ \Xhline{1pt}
\end{tabularx}
\label{tab:Precisionres512Appendix}
\end{table}

\begin{table}[tb!]
\centering
\small
\caption{The Precision (P) (higher is better) and Recall (R) (higher is better) scores of our method compared to state-of-the-art methods on \textbf{ \bm{$1024 \times 1024$} datasets with limited data amounts}.}
\begin{tabularx}{\textwidth}{c*{10}{|Y}}
\Xhline{1pt}

& \multicolumn{2}{c}{Shells}   & \multicolumn{2}{|c}{Skulls}      & \multicolumn{2}{|c}{Pokemon}     & \multicolumn{2}{|c}{BrecaHAD}      & \multicolumn{2}{|c}{MetFace}\\
& \multicolumn{2}{c}{64 imgs}   & \multicolumn{2}{|c}{97 imgs}      & \multicolumn{2}{|c}{833  imgs}     & \multicolumn{2}{|c}{162 imgs}      & \multicolumn{2}{|c}{1336 imgs}\\ \cline{2-11}
Method                                  & P            & R          & P          & R         & P         & R         & P        & R         & P        & R          \\ \Xhline{0.6pt}
StyleGAN2~\cite{karras2020analyzing}    & {\bf0.73}    & 0.03       & 0.12       & 0.02      & 0.69      & 0.00      & 0.53     & 0.01      & 0.74     & 0.00       \\
ADA~\cite{karras2020training}           & 0.52         & 0.03       & 0.72       & 0.03      & -         & -         & 0.82     & 0.12      & 0.78     & 0.23       \\
APA~\cite{jiang2021deceive}             & 0.50         & 0.02       & 0.76       & 0.06      & {\bf0.87} & 0.00      & 0.83     & 0.24      & 0.80     & 0.27       \\
DiffAug~\cite{DiffAug}                  & 0.56         & 0.00       & 0.57       & 0.00      & 0.69      & 0.01      & 0.68     & 0.04      & 0.82     & 0.24       \\
FastGAN~\cite{liu2021towards}           & 0.59         & 0.06       & 0.70       & 0.03      & 0.74      & 0.25      & {\bf0.94}& 0.42      & {\bf0.86}& 0.27       \\    \Xhline{0.4pt}
FreGAN (Ours)                           & 0.65         & {\bf0.08}  & {\bf0.83}  & {\bf0.08} & 0.80      & {\bf0.31} & {\bf0.94}& {\bf0.51} & {\bf0.86}& {\bf0.32}  \\ \Xhline{1pt}
\end{tabularx}
\label{tab:Precisionres1024Appendix}
\end{table}

\begin{table}[tb!]
\centering
\small
\caption{The  Precision (P) (higher is better) and Recall (R) (higher is better) scores of our method compared to the baseline FastGAN~\cite{liu2021towards} on \textbf{ AFHQ ($\sim$5k)~\cite{choi2020stargan} datasets with more data}.}
\begin{tabularx}{\textwidth}{c*{6}{|Y}}
\Xhline{1pt}
                                        & \multicolumn{2}{c}{AFHQ-Cat (5153 imgs)}   & \multicolumn{2}{|c}{AFHQ-Dog (4739 imgs)}    & \multicolumn{2}{|c}{AFHQ-Wild (4738 imgs)} \\ \cline{2-7}
Method                                  & P             & R                     & P            & R                        & P             & R           \\ \Xhline{0.6pt}
FastGAN~\cite{liu2021towards}           & 0.81          & 0.31                  & 0.86         & 0.56                     & 0.76          & {\bf0.22}   \\
+Ours                                   & {\bf0.82}     & {\bf0.45}             & {\bf0.87}    & {\bf0.69}                & {\bf0.77}     & 0.21        \\ \Xhline{1pt}
\end{tabularx}
\label{tab:PrecisionresMSAppendix}
\end{table}

\begin{table}[tb!]
\centering
\small
\caption{The Density (D) (higher is better) and Coverage (C) (higher is better) scores of our method compared to state-of-the-art methods on \textbf{ \bm{$256 \times 256$} datasets with limited data amounts}.}
\begin{tabularx}{\textwidth}{c*{10}{|Y}}
\Xhline{1pt}
                                        & \multicolumn{4}{c|}{{Animal Face}}                                            & \multicolumn{6}{c}{{100-shot}} \\ \cline{2-11}
                                        & \multicolumn{2}{c}{Dog (389 imgs)}   & \multicolumn{2}{|c}{Cat (160 imgs)}      & \multicolumn{2}{|c}{Panda}     & \multicolumn{2}{|c}{Obama}      & \multicolumn{2}{|c}{Grumpy\_cat}\\ \cline{2-11}
Method                                  & D            & C          & D          & C         & D         & C         & D        & C         & D         & C          \\ \Xhline{0.6pt}
StyleGAN2~\cite{karras2020analyzing}    & 0.19         & 0.51       & 0.28       & 0.89      & 0.06      & 0.99      & 0.26     & 0.98      & 0.59      & 0.99       \\
ADA~\cite{karras2020training}           & 0.61         & 0.96       & 0.85       & {\bf1.00} & 0.06      & {\bf1.00} & 1.23     & {\bf1.00} & 0.30      & 0.45       \\
APA~\cite{jiang2021deceive}             & 0.62         & 0.74       & 0.78       & 0.99      & 0.14      & 0.91      & 0.97     & {\bf1.00} & 0.95      & {\bf1.00}  \\
DiffAug~\cite{DiffAug}                  & 0.65         & 0.91       & 1.13       & {\bf1.00} & 0.11      & {\bf1.00} & 0.68     & {\bf1.00} & {\bf1.37} & {\bf1.00}  \\
FastGAN~\cite{liu2021towards}           & {\bf0.87}    & 0.96       & 1.06       & {\bf1.00} & 0.10      & {\bf1.00} & 1.28     & {\bf1.00} & 1.30      & {\bf1.00}  \\    \Xhline{0.4pt}
FreGAN (Ours)                           & 0.86         & {\bf0.98}  & {\bf1.24}  & {\bf1.00} & {\bf0.15} & {\bf1.00} & {\bf1.38}& {\bf1.00} & 1.28      & {\bf1.00}  \\ \Xhline{1pt}
\end{tabularx}
\label{tab:Diversity256Appendix1}
\end{table}

\begin{table}[tb!]
\centering
\small
\caption{The Density (D) (higher is better) and Coverage (C) (higher is better) scores of our method compared to state-of-the-art methods on \textbf{ \bm{$256 \times 256$} datasets with limited data amounts}.}
\begin{tabularx}{\textwidth}{c*{8}{|Y}}
\Xhline{1pt}
                                        & \multicolumn{2}{c}{Medici (100 imgs)}& \multicolumn{2}{|c}{Temple (100 imgs)} &\multicolumn{2}{|c}{Bridge (100 imgs)} & \multicolumn{2}{|c}{Wuzhen (100 imgs)}  \\ \cline{2-9}
Method                                  & D            & C           & D          & C         & D          & C         & D         & C        \\ \Xhline{0.6pt}
StyleGAN2~\cite{karras2020analyzing}    & 0.11         & 0.79        & 0.31       & 0.81      & 0.64       & 0.55      & 0.04      & 0.67     \\
ADA~\cite{karras2020training}           & -            & -           & -          & -         & -          & -         & 0.27      & 0.90     \\
APA~\cite{jiang2021deceive}             & -            & -           & {\bf1.27}  & 0.98      & 0.06       & 0.23      & 0.29      & 0.96     \\
DiffAug~\cite{DiffAug}                  & 0.35         & 0.86        & 0.91       & {\bf1.00} & 1.03       & {\bf1.00} & 0.14      & 0.83     \\
FastGAN~\cite{liu2021towards}           & 0.88         & 0.98        & 1.19       & {\bf1.00} & 1.04       & {\bf1.00} & 0.94      & {\bf1.00}\\ \Xhline{0.4pt}
FreGAN (Ours)                           & {\bf1.00}    & {\bf1.00}   & 1.24      & {\bf1.00} & {\bf1.16}  & {\bf1.00} & {\bf1.20} & {\bf1.00}\\ \Xhline{1pt}
\end{tabularx}
\label{tab:res256DiversityAppendix2}
\end{table}

\begin{table}[tb!]
\centering
\small
\caption{The Density (D) (higher is better) and Coverage (C) (higher is better) scores of our method compared to state-of-the-art methods on \textbf{ \bm{$512 \times 512$} datasets with limited data amounts}.}
\begin{tabularx}{\textwidth}{c*{10}{|Y}}
\Xhline{1pt}
                            & \multicolumn{2}{c}{AnimeFace}   & \multicolumn{2}{|c}{ArtPainting}      & \multicolumn{2}{|c}{Moongate}     & \multicolumn{2}{|c}{Flat}      & \multicolumn{2}{|c}{Fauvism}\\
 & \multicolumn{2}{c}{120 imgs}   & \multicolumn{2}{|c}{1000 imgs}      & \multicolumn{2}{|c}{136 imgs}     & \multicolumn{2}{|c}{36 imgs}     & \multicolumn{2}{|c}{124 imgs}\\ \cline{2-11}
Method                                  & D             & C          & D          & C         & D         & C         & D        & C         & D         & C     \\ \Xhline{0.6pt}
StyleGAN2~\cite{karras2020analyzing}    & 0.25          & 0.10       & 0.16       & 0.34      & 0.19      & 0.38      & 0.36     & 0.67      & 0.16      & 0.27  \\
ADA~\cite{karras2020training}           & 1.58          & 0.99       & 0.93       & 0.91      & 0.45      & 0.99      & 0.90     & 0.83      & 0.99      & 0.90  \\
APA~\cite{jiang2021deceive}             & {\bf2.15}     & {\bf1.00}  & 0.72       & 0.90      & 0.42      & 0.96      & 0.37     & 0.67      & 1.13      & 0.93  \\
DiffAug~\cite{DiffAug}                  & 0.22          & 0.30       & 0.90       & 0.87      & 0.86      & 0.90      & 0.93     & 0.56      & 0.32      & 0.75  \\
FastGAN~\cite{liu2021towards}           & 1.27          & {\bf1.00}  & 1.17       & 0.95      & 0.93      & {\bf1.00} & 0.74     & 0.97      & 1.53      & 0.99  \\    \Xhline{0.4pt}
FreGAN (Ours)                           & 1.65          & {\bf1.00}  & {\bf1.23}  & {\bf0.97} & {\bf1.17} & {\bf1.00} & {\bf0.94}& {\bf1.00} & {\bf1.33} & {\bf1.00} \\ \Xhline{1pt}
\end{tabularx}
\label{tab:res512DiversityAppendix}
\end{table}

\begin{table}[tb!]
\centering
\small
\caption{The Density (D) (higher is better) and Coverage (C) (higher is better) scores of our method compared to state-of-the-art methods on \textbf{ \bm{$1024 \times 1024$} datasets with limited data amounts}.}
\begin{tabularx}{\textwidth}{c*{10}{|Y}}
\Xhline{1pt}

& \multicolumn{2}{c}{Shells}   & \multicolumn{2}{|c}{Skulls}      & \multicolumn{2}{|c}{Pokemon}     & \multicolumn{2}{|c}{BrecaHAD}      & \multicolumn{2}{|c}{MetFace}\\
& \multicolumn{2}{c}{64 imgs}   & \multicolumn{2}{|c}{97 imgs}      & \multicolumn{2}{|c}{833  imgs}     & \multicolumn{2}{|c}{162 imgs}      & \multicolumn{2}{|c}{1336 imgs}\\ \cline{2-11}
Method                                  & D            & C          & D          & C         & D         & C         & D        & C         & D        & C         \\ \Xhline{0.6pt}
StyleGAN2~\cite{karras2020analyzing}    & {\bf0.97}    & {\bf1.00}  & 0.11       & 0.71      & 0.39      & 0.13      & 0.23     & 0.47      & 0.61     & 0.51       \\
ADA~\cite{karras2020training}           & 0.43         & {\bf1.00}  & 0.86       & {\bf1.00} & -         & -         & 0.75     & {\bf1.00} & 0.93     & 0.96       \\
APA~\cite{jiang2021deceive}             & 0.35         & 0.89       & 0.91       & {\bf1.00} & {\bf1.22} & 0.71      & 0.68     & {\bf1.00} & 1.12     & 0.96       \\
DiffAug~\cite{DiffAug}                  & 0.47         & 0.94       & 0.58       & 0.99      & 0.50      & 0.63      & 0.52     & 0.96      & 1.08     & 0.96       \\
FastGAN~\cite{liu2021towards}           & 0.66         & 0.94       & 1.34       & {\bf1.00} & 0.92      & 0.96      & {\bf1.20}& {\bf1.00} & {\bf1.37}& 0.96       \\    \Xhline{0.4pt}
FreGAN (Ours)                           & 0.78         & {\bf1.00}  & {\bf1.35}  & {\bf1.00} & 1.10      & {\bf0.97} & 1.08     & {\bf1.00} & 1.27     & {\bf0.97}  \\ \Xhline{1pt}
\end{tabularx}
\label{tab:res1024DiversityAppendix}
\end{table}

\begin{table}[tb!]
\centering
\small
\caption{The Density (D) (higher is better) and Coverage (C) (higher is better) scores of our method compared to the baseline FastGAN~\cite{liu2021towards} on \textbf{ AFHQ ($\sim$5k)~\cite{choi2020stargan} datasets with more data}.}
\begin{tabularx}{\textwidth}{c*{6}{|Y}}
\Xhline{1pt}
                                        & \multicolumn{2}{c}{AFHQ-Cat (5153 imgs)}   & \multicolumn{2}{|c}{AFHQ-Dog (4739 imgs)}    & \multicolumn{2}{|c}{AFHQ-Wild (4738 imgs)} \\ \cline{2-7}
Method                                  & D             & C                     & D            & C                        & D             & C           \\ \Xhline{0.6pt}
FastGAN~\cite{liu2021towards}           & {\bf1.19}     & 0.80                  & {\bf0.82}    & 0.50                     & 1.21          & 0.72        \\
+Ours                                   & 1.18          & {\bf0.85}             & 0.72         & {\bf0.57}                & {\bf1.24}     & {\bf0.73}   \\ \Xhline{1pt}
\end{tabularx}
\label{tab:DensityresMSAppendix}
\end{table}

\begin{table}[tb!]
\centering
\small
\caption{The IS (higher is better) scores of our method compared to state-of-the-art methods on \textbf{ \bm{$256 \times 256$} datasets with limited data amounts}.}
\begin{tabularx}{\textwidth}{c*{5}{|Y}}
\Xhline{1pt}
                                & \multicolumn{2}{c|}{{Animal Face}}                                            & \multicolumn{3}{c}{{100-shot}} \\ \cline{2-6}
& \multicolumn{1}{c}{Dog (389 imgs)}   & \multicolumn{1}{|c}{Cat (160 imgs)}      & \multicolumn{1}{|c}{Panda}     & \multicolumn{1}{|c}{Obama}      & \multicolumn{1}{|c}{Grumpy\_cat}\\ \cline{2-6}
Method                                  & IS                & IS              & IS              & IS                & IS            \\ \Xhline{0.6pt}
StyleGAN2~\cite{karras2020analyzing}    & 7.29$\pm$0.33         & 2.37$\pm$0.08       & {\bf1.03}$\pm$0.01  & {\bf1.67}$\pm$0.05    & 1.31$\pm$0.02     \\
ADA~\cite{karras2020training}           & 8.13$\pm$0.30         & 2.43$\pm$0.06       & 1.01$\pm$0.00       & 1.38$\pm$0.03         & 1.10$\pm$0.01     \\
APA~\cite{jiang2021deceive}             & 7.35$\pm$0.27         & 2.37$\pm$0.08       & 1.02$\pm$0.00       & 1.45$\pm$0.01         & {\bf1.43}$\pm$0.02\\
DiffAug~\cite{DiffAug}                  & 8.22$\pm$0.31         & 2.03$\pm$0.05       & 1.01$\pm$0.00       & 1.29$\pm$0.02         & 1.29$\pm$0.01     \\
FastGAN~\cite{liu2021towards}           & 7.60$\pm$0.30         & 2.28$\pm$0.06       & 1.00$\pm$0.00       & 1.32$\pm$0.02         & 1.33$\pm$0.02     \\    \Xhline{0.4pt}
FreGAN (Ours)                           & {\bf8.75}$\pm$0.33    & {\bf2.47}$\pm$0.06  & 1.00$\pm$0.00       & 1.50$\pm$0.02         & 1.35$\pm$0.01     \\ \Xhline{1pt}
\end{tabularx}
\label{tab:res256ISAppendix1}
\end{table}

\begin{table}[tb!]
\centering
\small
\caption{The IS (higher is better) scores of our method compared to state-of-the-art methods on \textbf{ \bm{$256 \times 256$} datasets with limited data amounts}.}
\begin{tabularx}{\textwidth}{c*{4}{|Y}}
\Xhline{1pt}
                                        & \multicolumn{1}{c}{Medici (100 imgs)}& \multicolumn{1}{|c}{Temple (100 imgs)} &\multicolumn{1}{|c}{Bridge (100 imgs)} & \multicolumn{1}{|c}{Wuzhen (100 imgs)}  \\ \cline{2-5}
Method                                  & IS                & IS                & IS                & IS                 \\ \Xhline{0.6pt}
StyleGAN2~\cite{karras2020analyzing}    & 1.32$\pm$0.03         & {\bf2.04}$\pm$0.06    & 1.60$\pm$0.03         & 1.93$\pm$0.04          \\
ADA~\cite{karras2020training}           & {\bf2.18}$\pm$0.04    & 1.98$\pm$0.04         & {\bf1.92}$\pm$0.03    & 2.02$\pm$0.03          \\
APA~\cite{jiang2021deceive}             & 1.24$\pm$0.01         & 1.60$\pm$0.02         & 1.08$\pm$0.01         & {\bf2.43}$\pm$0.07     \\
DiffAug~\cite{DiffAug}                  & 1.78$\pm$0.03         & 1.76$\pm$0.02         & 1.68$\pm$0.03         & 1.93$\pm$0.05          \\
FastGAN~\cite{liu2021towards}           & 1.76$\pm$0.04         & 1.63$\pm$0.02         & 1.55$\pm$0.02         & 1.99$\pm$0.04          \\ \Xhline{0.4pt}
FreGAN (Ours)                           & 1.69$\pm$0.02         & 1.62$\pm$0.01         & 1.59$\pm$0.03         & 2.12$\pm$0.05          \\ \Xhline{1pt}
\end{tabularx}
\label{tab:res256ISAppendix2}
\end{table}

\begin{table}[tb!]
\centering
\small
\caption{The IS (higher is better) scores of our method compared to state-of-the-art methods on \textbf{ \bm{$512 \times 512$} datasets with limited data amounts}.}
\begin{tabularx}{\textwidth}{c*{5}{|Y}}
\Xhline{1pt}
                            & \multicolumn{1}{c}{AnimeFace}  & \multicolumn{1}{|c}{ArtPainting}      & \multicolumn{1}{|c}{Moongate}     & \multicolumn{1}{|c}{Flat}      & \multicolumn{1}{|c}{Fauvism}\\
                            & \multicolumn{1}{c}{120 imgs}        & \multicolumn{1}{|c}{1000 imgs}      & \multicolumn{1}{|c}{136 imgs}     & \multicolumn{1}{|c}{36 imgs}      & \multicolumn{1}{|c}{124 imgs}\\ \cline{2-6}
Method                                  & IS                & IS              & IS              & IS             & IS               \\ \Xhline{0.6pt}
StyleGAN2~\cite{karras2020analyzing}    & 1.37$\pm$0.02         & 2.79$\pm$0.08       & 3.82$\pm$0.06       & 2.21$\pm$0.04      & 2.32$\pm$0.04        \\
ADA~\cite{karras2020training}           & 1.93$\pm$0.04         & 3.64$\pm$0.12       & 4.12$\pm$0.19       & 3.77$\pm$0.09      & 2.93$\pm$0.07        \\
APA~\cite{jiang2021deceive}             & 1.85$\pm$0.06         & {\bf4.65}$\pm$0.17  & {\bf4.19}$\pm$0.17  & 3.02$\pm$0.05      & {\bf3.68}$\pm$0.10   \\
DiffAug~\cite{DiffAug}                  & 1.19$\pm$0.01         & 3.39$\pm$0.06       & 2.66$\pm$0.07       & 1.48$\pm$0.02      & 3.14$\pm$0.10        \\
FastGAN~\cite{liu2021towards}           & 2.06$\pm$0.04         & 4.25$\pm$0.12       & 2.96$\pm$0.09       & 4.55$\pm$0.14      & 3.24$\pm$0.07        \\ \Xhline{0.4pt}
FreGAN (Ours)                           & {\bf2.08}$\pm$0.03    & 4.17$\pm$0.14       & 3.39$\pm$0.07       & {\bf5.18}$\pm$0.21 & 3.20$\pm$0.10        \\ \Xhline{1pt}
\end{tabularx}
\label{tab:res512ISAppendix}
\end{table}

\begin{table}[tb!]
\centering
\small
\caption{The IS (higher is better) scores of our method compared to state-of-the-art methods on \textbf{ \bm{$1024 \times 1024$} datasets with limited data amounts}.}
\begin{tabularx}{\textwidth}{c*{5}{|Y}}
\Xhline{1pt}
& \multicolumn{1}{c}{Shells}   & \multicolumn{1}{|c}{Skulls}      & \multicolumn{1}{|c}{Pokemon}     & \multicolumn{1}{|c}{BrecaHAD}      & \multicolumn{1}{|c}{MetFace}\\
& \multicolumn{1}{c}{64 imgs}   & \multicolumn{1}{|c}{97 imgs}      & \multicolumn{1}{|c}{833 imgs}     & \multicolumn{1}{|c}{162 imgs}      & \multicolumn{1}{|c}{1336 imgs}\\ \cline{2-6}
Method                                  & IS                & IS              & IS              & IS             & IS               \\ \Xhline{0.6pt}
StyleGAN2~\cite{karras2020analyzing}    & 3.24$\pm$0.10         & {\bf3.97}$\pm$0.09  & 2.13$\pm$0.04       & 1.65$\pm$0.03      & 1.92$\pm$0.04        \\
ADA~\cite{karras2020training}           & 3.78$\pm$0.09         & 2.61$\pm$0.08       & 1.41$\pm$0.02       & 2.82$\pm$0.10      & 3.38$\pm$0.09        \\
APA~\cite{jiang2021deceive}             & {\bf3.83}$\pm$0.14    & 2.82$\pm$0.09       & 1.99$\pm$0.44       & {\bf3.01}$\pm$0.13 & {\bf3.51}$\pm$0.12   \\
DiffAug~\cite{DiffAug}                  & 3.14$\pm$0.10         & 2.67$\pm$0.08       & 2.60$\pm$0.07       & 2.72$\pm$0.10      & 3.29$\pm$0.10        \\
FastGAN~\cite{liu2021towards}           & 2.52$\pm$0.09         & 2.24$\pm$0.08       & {\bf2.63}$\pm$0.09  & 2.83$\pm$0.04      & 3.07$\pm$0.08        \\    \Xhline{0.4pt}
FreGAN (Ours)                           & 2.72$\pm$0.06         & 2.47$\pm$0.06       & 2.38$\pm$0.03       & {\bf3.01}$\pm$0.05 & 3.06$\pm$0.08        \\ \Xhline{1pt}
\end{tabularx}
\label{tab:res1024ISAppendix}
\end{table}

\begin{table}[tb!]
\centering
\small
\caption{The IS (higher is better) scores of our method compared to the baseline FastGAN~\cite{liu2021towards} on \textbf{ AFHQ ($\sim$5k)~\cite{choi2020stargan} datasets with more data}.}
\begin{tabularx}{\textwidth}{c*{3}{|Y}}
\Xhline{1pt}
                                        & \multicolumn{1}{c}{AFHQ-Cat (5153 imgs)}   & \multicolumn{1}{|c}{AFHQ-Dog (4739 imgs)}    & \multicolumn{1}{|c}{AFHQ-Wild (4738 imgs)} \\ \cline{2-4}
Method                                  & IS                        & IS                     & IS                       \\ \Xhline{0.6pt}
FastGAN~\cite{liu2021towards}           & 1.93$\pm$0.02             & 8.44$\pm$0.27          & 5.08$\pm$0.08            \\
+Ours                                   & {\bf2.07}$\pm$0.05        & {\bf9.23}$\pm$0.31     & {\bf5.14}$\pm$0.10       \\ \Xhline{1pt}
\end{tabularx}
\label{tab:ISresMSAppendix}
\end{table}

\newpage
\begin{table}[tb!]
\centering
\small
\caption{{The LPIPS (higher is better) scores of our method compared to state-of-the-art methods on  $256 \times 256$ datasets with limited data amounts.}}
\begin{tabularx}{\textwidth}{c*{5}{|Y}}
\Xhline{1pt}
                                & \multicolumn{2}{c|}{{Animal Face}}                                            & \multicolumn{3}{c}{{100-shot}} \\ \cline{2-6}
& \multicolumn{1}{c}{Dog (389 imgs)}   & \multicolumn{1}{|c}{Cat (160 imgs)}      & \multicolumn{1}{|c}{Panda}     & \multicolumn{1}{|c}{Obama}      & \multicolumn{1}{|c}{Grumpy\_cat}\\ \cline{2-6}
Method                                  & LPIPS                     & LPIPS                   & LPIPS                   & LPIPS                     & LPIPS            \\ \Xhline{0.6pt}
StyleGAN2~\cite{karras2020analyzing}    & 0.6550$\pm$0.0011         & 0.5774$\pm$0.0010       & 0.4935$\pm$0.0013       & 0.5264$\pm$0.0013         & 0.4632$\pm$0.0013     \\
ADA~\cite{karras2020training}           & 0.6499$\pm$0.0011         & 0.6145$\pm$0.0014       & 0.4996$\pm$0.0012       & 0.4717$\pm$0.0015         & 0.4721$\pm$0.0017     \\
APA~\cite{jiang2021deceive}             & 0.6296$\pm$0.0013         & 0.6310$\pm$0.0013       & 0.5037$\pm$0.0011       & 0.4922$\pm$0.0012         & 0.4519$\pm$0.0011     \\
DiffAug~\cite{DiffAug}                  & 0.6301$\pm$0.0009         & 0.5453$\pm$0.0013       & 0.5144$\pm$0.0010       & 0.4727$\pm$0.0013         & 0.4426$\pm$0.0013     \\
FastGAN~\cite{liu2021towards}           & 0.6751$\pm$0.0011         & 0.6452$\pm$0.0014       & 0.6073$\pm$0.0009       & {\bf0.6081}$\pm$0.0012     & 0.6077$\pm$0.0009     \\    \Xhline{0.4pt}
FreGAN (Ours)                           & {\bf0.6848}$\pm$0.0011    & {\bf0.6671}$\pm$0.0014  & {\bf0.6089}$\pm$0.0010  & 0.6025$\pm$0.0011         & {\bf0.6149}$\pm$0.0009     \\ \Xhline{1pt}
\end{tabularx}
\label{tab:res256LPIPS1}
\end{table}

\begin{table}[tb!]
\centering
\small
\caption{{The LPIPS (higher is better) scores of our method compared to state-of-the-art methods on $256 \times 256$ datasets with limited data amounts.}}
\begin{tabularx}{\textwidth}{c*{4}{|Y}}
\Xhline{1pt}
                                        & \multicolumn{1}{c}{Medici (100 imgs)}& \multicolumn{1}{|c}{Temple (100 imgs)} &\multicolumn{1}{|c}{Bridge (100 imgs)} & \multicolumn{1}{|c}{Wuzhen (100 imgs)}  \\ \cline{2-5}
Method                                  & LPIPS                     & LPIPS                     & LPIPS                     & LPIPS                           \\ \Xhline{0.6pt}
StyleGAN2~\cite{karras2020analyzing}    & 0.5410$\pm$0.0013         & 0.5280$\pm$0.0017         & 0.4651$\pm$0.0014         & {\bf0.6736}$\pm$0.0012          \\
ADA~\cite{karras2020training}           & 0.5182$\pm$0.0030         & 0.4389$\pm$0.0024         & 0.5341$\pm$0.0021         & 0.6086$\pm$0.0015               \\
APA~\cite{jiang2021deceive}             & 0.3550$\pm$0.0027         & 0.4909$\pm$0.0017         & 0.5872$\pm$0.0013         & 0.6622$\pm$0.0014               \\
DiffAug~\cite{DiffAug}                  & 0.4683$\pm$0.0017         & 0.5250$\pm$0.0014         & 0.5674$\pm$0.0011         & 0.6724$\pm$0.0013               \\
FastGAN~\cite{liu2021towards}           & {\bf0.5298}$\pm$0.0024    & 0.5229$\pm$0.0015         & {\bf0.5958}$\pm$0.0010    & 0.6630$\pm$0.0011               \\ \Xhline{0.4pt}
FreGAN (Ours)                           & {\bf0.5298}$\pm$0.0024    & {\bf0.5275}$\pm$0.0015         & 0.5947$\pm$0.0009         & 0.6656$\pm$0.0011               \\ \Xhline{1pt}
\end{tabularx}
\label{tab:res256LPIPS2}
\end{table}

\begin{table}[tb!]
\centering
\small
\caption{{The LPIPS (higher is better) scores of our method compared to state-of-the-art methods on $512 \times 512$ datasets with limited data amounts.}}
\begin{tabularx}{\textwidth}{c*{5}{|Y}}
\Xhline{1pt}
                            & \multicolumn{1}{c}{AnimeFace}  & \multicolumn{1}{|c}{ArtPainting}      & \multicolumn{1}{|c}{Moongate}     & \multicolumn{1}{|c}{Flat}      & \multicolumn{1}{|c}{Fauvism}\\
                            & \multicolumn{1}{c}{120 imgs}        & \multicolumn{1}{|c}{1000 imgs}      & \multicolumn{1}{|c}{136 imgs}     & \multicolumn{1}{|c}{36 imgs}      & \multicolumn{1}{|c}{124 imgs}\\ \cline{2-6}
Method                                  & LPIPS                     & LPIPS                   & LPIPS                   & LPIPS                  & LPIPS                    \\ \Xhline{0.6pt}
StyleGAN2~\cite{karras2020analyzing}    & 0.4253$\pm$0.0020         & 0.7244$\pm$0.0009       & 0.7047$\pm$0.0026       & 0.6223$\pm$0.0026      & 0.6344$\pm$0.0009        \\
ADA~\cite{karras2020training}           & 0.5611$\pm$0.0015         & 0.8102$\pm$0.0015       & 0.6418$\pm$0.0015       & 0.7288$\pm$0.0017      & 0.6509$\pm$0.0014        \\
APA~\cite{jiang2021deceive}             & 0.5491$\pm$0.0017         & 0.8062$\pm$0.0014       & {\bf0.7235}$\pm$0.0016  & 0.6317$\pm$0.0022      & 0.6848$\pm$0.0014        \\
DiffAug~\cite{DiffAug}                  & 0.4926$\pm$0.0005         & 0.7717$\pm$0.0016       & 0.5880$\pm$0.0015       & 0.4403$\pm$0.0005      & 0.6117$\pm$0.0023        \\
FastGAN~\cite{liu2021towards}           & 0.6188$\pm$0.0011         & 0.8344$\pm$0.0015       & 0.6603$\pm$0.0010       & 0.7939$\pm$0.0016      & {\bf0.7028}$\pm$0.0010   \\ \Xhline{0.4pt}
FreGAN (Ours)                           & {\bf0.6191}$\pm$0.0010    & {\bf0.8439}$\pm$0.0016  & 0.6673$\pm$0.0016       & {\bf0.7952}$\pm$0.0011 & {\bf0.7028}$\pm$0.0010   \\ \Xhline{1pt}
\end{tabularx}
\label{tab:res512LPIPS}
\end{table}

\begin{table}[tb!]
\centering
\small
\caption{{The LPIPS (higher is better) scores of our method compared to state-of-the-art methods on $1024 \times 1024$ datasets with limited data amounts.}}
\begin{tabularx}{\textwidth}{c*{5}{|Y}}
\Xhline{1pt}
& \multicolumn{1}{c}{Shells}   & \multicolumn{1}{|c}{Skulls}      & \multicolumn{1}{|c}{Pokemon}     & \multicolumn{1}{|c}{BrecaHAD}      & \multicolumn{1}{|c}{MetFace}\\
& \multicolumn{1}{c}{64 imgs}   & \multicolumn{1}{|c}{97 imgs}      & \multicolumn{1}{|c}{833 imgs}     & \multicolumn{1}{|c}{162 imgs}      & \multicolumn{1}{|c}{1336 imgs}\\ \cline{2-6}
Method                                  & LPIPS                     & LPIPS                   & LPIPS                   & LPIPS                  & LPIPS               \\ \Xhline{0.6pt}
StyleGAN2~\cite{karras2020analyzing}    & {\bf0.5486}$\pm$0.0018    & {\bf0.6565}$\pm$0.0024  & 0.5870$\pm$0.0006       & 0.4604$\pm$0.0016      & 0.5321$\pm$0.0016        \\
ADA~\cite{karras2020training}           & 0.5268$\pm$0.0015         & 0.5857$\pm$0.0025       & 0.4050$\pm$0.0015       & 0.5524$\pm$0.0016      & 0.6648$\pm$0.0013        \\
APA~\cite{jiang2021deceive}             & 0.5337$\pm$0.0017         & 0.6208$\pm$0.0025       & 0.4241$\pm$0.0013       & 0.4874$\pm$0.0016      & {\bf0.6947}$\pm$0.0014   \\
DiffAug~\cite{DiffAug}                  & 0.4935$\pm$0.0015         & 0.6027$\pm$0.0028       & 0.4811$\pm$0.0011       & 0.4961$\pm$0.0024      & 0.6579$\pm$0.0013        \\
FastGAN~\cite{liu2021towards}           & 0.4908$\pm$0.0012         & 0.6115$\pm$0.0028       & 0.5705$\pm$0.0008       & {\bf0.5597}$\pm$0.0012 & 0.6698$\pm$0.0013        \\    \Xhline{0.4pt}
FreGAN (Ours)                           & 0.4957$\pm$0.0014         & 0.6112$\pm$0.0026       & {\bf0.5770}$\pm$0.0008  & 0.5582$\pm$0.0011      & 0.6706$\pm$0.0012        \\ \Xhline{1pt}
\end{tabularx}
\label{tab:res1024LPIPS}
\end{table}

\newpage

\section{Results of Large Datasets}
\label{sec:largedatasets}
{We evaluate the effectiveness of our method on large-scale datasets in Tab.~\ref{tab:CelebAresults} and Tab.~\ref{tab:FFHQresults}.
Specifically, we use the whole datasets of celebA~\cite{liu2015deep} as training data, and we randomly select 30k images from FFHQ~\cite{karras2019style} as training data.
We generate 50k images for quantitative evaluation, we can observe from Tab.~\ref{tab:CelebAresults} and Tab.~\ref{tab:FFHQresults} that our FreGAN also improves the generation quality on large-scale training data.
In the future, we will also combine our proposed method with large-scale GANs like StyleGAN3~\cite{Karras2021} and BigGAN~\cite{donahue2019large} for more comprehensive investigation.}

\begin{table}[tb!]
\centering
\small
\caption{{Comparison results of our method compared to the baseline FastGAN~\cite{liu2021towards} on \textbf{CelebA dataset with 30k training images}.}}
\begin{tabularx}{\textwidth}{c*{7}{|Y}}
\Xhline{1pt}
Method                              & FID ($\downarrow$)  & KID ($\downarrow$)  & IS                    & Precision                & Recall        & Density     & Coverage       \\ \Xhline{0.6pt}
FastGAN~\cite{liu2021towards}           & 23.35           & 19.04               & 2.57$\pm$0.07         & 0.69                     & 0.25          & 0.60        & 0.35           \\
FreGAN (Ours)                           & {\bf20.65}      & {\bf15.73}          & {\bf2.58}$\pm$0.06    & {\bf0.70}                & {\bf0.29}     & {\bf0.67}   & {\bf0.39}      \\ \Xhline{1pt}
\end{tabularx}
\label{tab:CelebAresults}
\end{table}

\begin{table}[tb!]
\centering
\small
\caption{{Comparison results of our method compared to the baseline FastGAN~\cite{liu2021towards} on \textbf{FFHQ dataset with 30k training images}.}}
\begin{tabularx}{\textwidth}{c*{7}{|Y}}
\Xhline{1pt}
Method                              & FID ($\downarrow$)  & KID ($\downarrow$)  & IS                    & Precision                & Recall        & Density     & Coverage       \\ \Xhline{0.6pt}
FastGAN~\cite{liu2021towards}       & 26.17               & 14.77               & 3.41$\pm$0.15         & 0.67                     & 0.34          & 0.61        & 0.42           \\
FreGAN (Ours)                       & {\bf23.62}          & {\bf12.44}          & {\bf3.51}$\pm$0.10    & {\bf0.69}                & {\bf0.37}     & {\bf0.66}   & {\bf0.44}      \\ \Xhline{1pt}
\end{tabularx}
\label{tab:FFHQresults}
\end{table}

\subsection{More Qualitative Results}
\label{sec:qualitativeapp}

We provide more qualitative results of FastGAN and our FreGAN in Fig.~\ref{fig:Vis256Appendix}, Fig.~\ref{fig:Vis512Appendix}, Fig.~\ref{fig:Vis1024Appendix}, respectively.
Our FreGAN is capable of generating more realistic images with more fine details.
The high-frequency components of our generated images contain more information, such as the eyes and mouth of the Panda in Fig.~\ref{fig:Vis256Appendix}, the texture of Moongate in Fig.~\ref{fig:Vis512Appendix}.

\begin{figure}
	\centering
	\vspace{-0.1cm}
	\includegraphics[width=\linewidth]{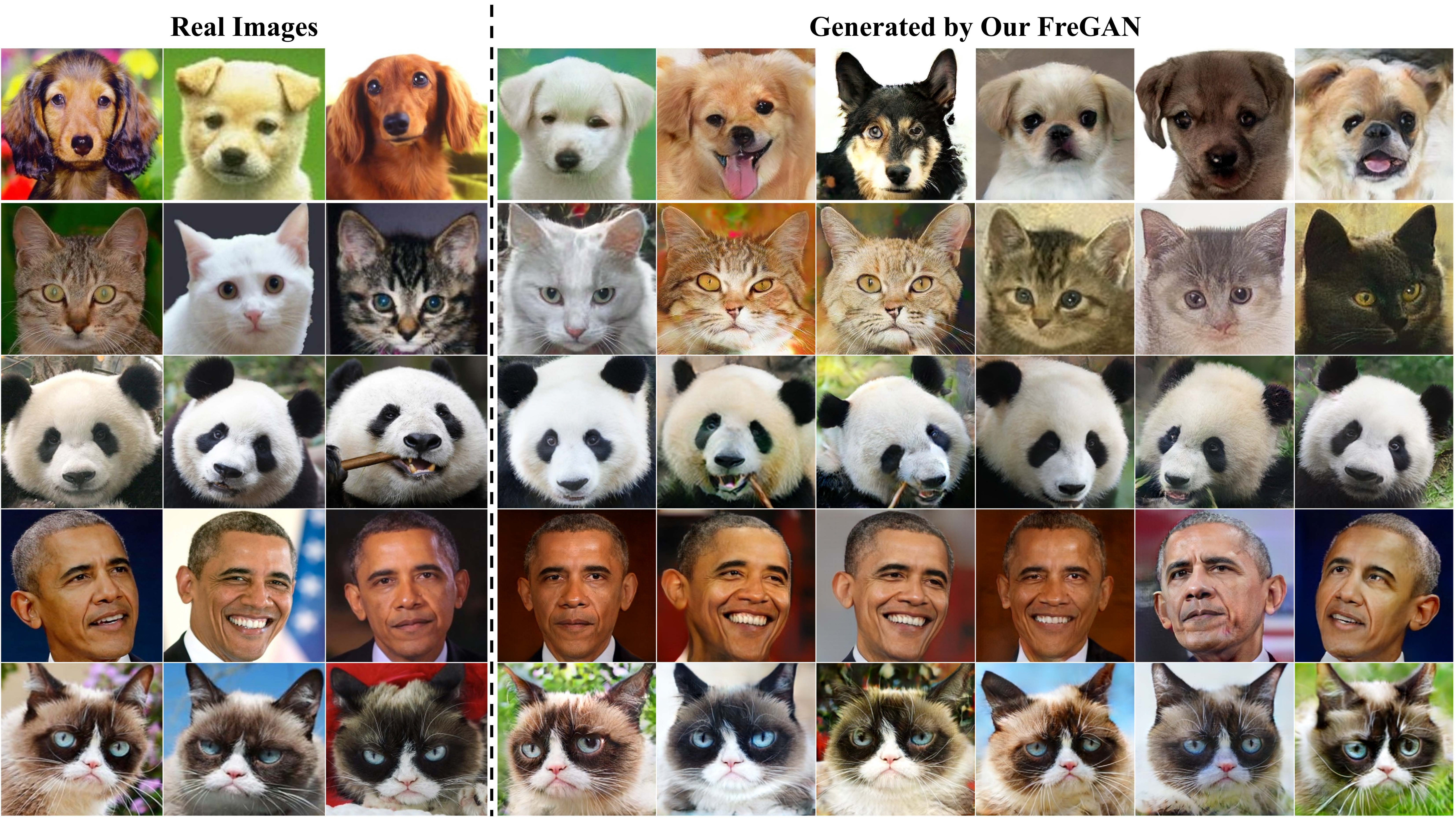}
	\vspace{-0.4cm}
	\caption{\textbf{Qualitative results of our FreGAN.} The left part shows some of real training images and the right part of images are generated by our FreGAN.
    Our FreGAN is capable of generating photorealistic images with fine details, which is indistinguishable by the discriminator.}
	\label{fig:Generated256Appendix}
	\vspace{-3mm}
\end{figure}

\begin{figure}
	\centering
	\vspace{-0.1cm}
	\includegraphics[width=\linewidth]{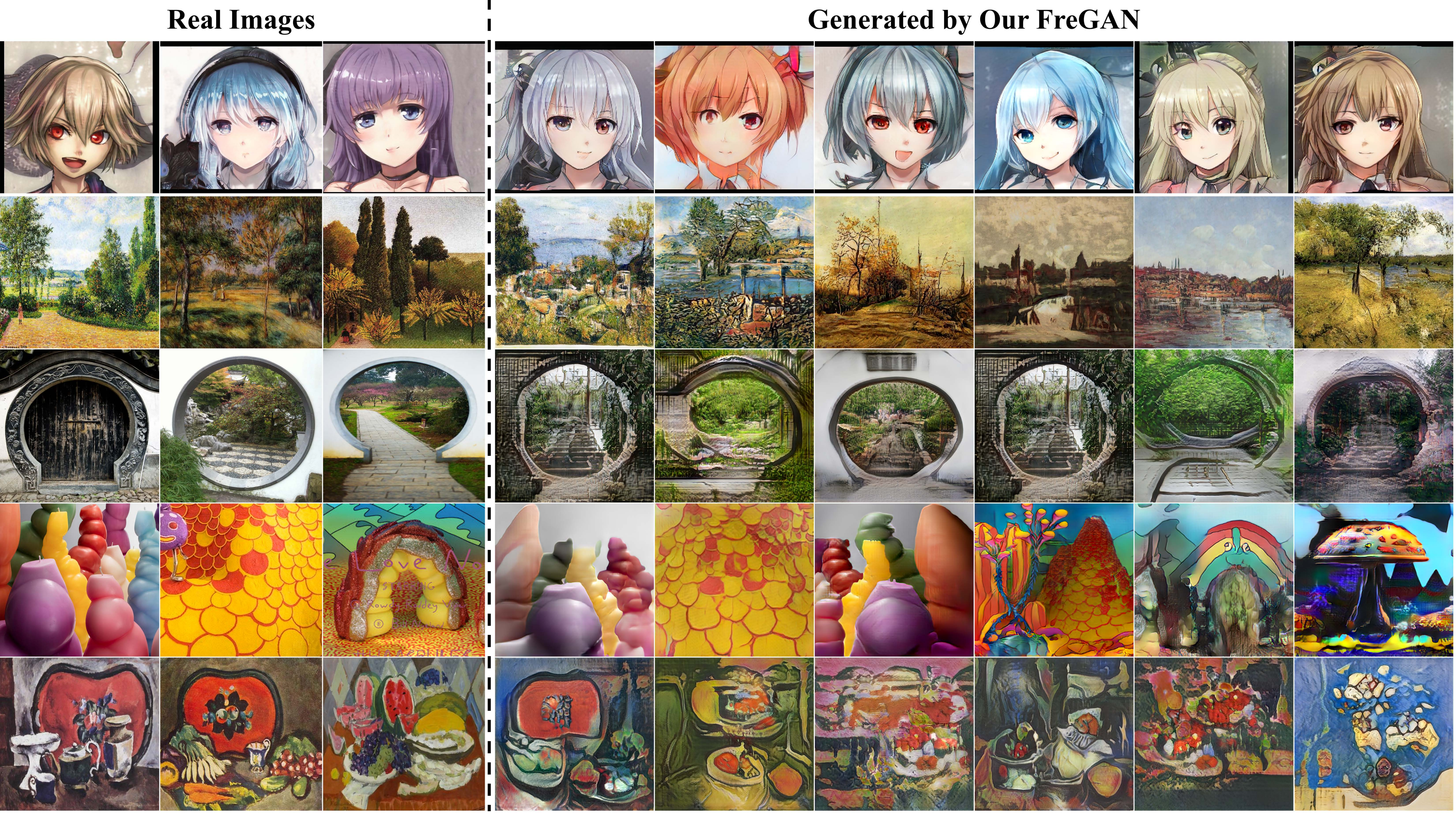}
	\vspace{-0.4cm}
    \caption{\textbf{Qualitative results of our FreGAN.} The left part shows some of real training images and the right part of images are generated by our FreGAN.
    Our FreGAN is capable of generating photorealistic images with fine details, which is indistinguishable by the discriminator.}
	\label{fig:Generated512Appendix}
	\vspace{-3mm}
\end{figure}

\begin{figure}
	\centering
	\vspace{-0.1cm}
	\includegraphics[width=\linewidth]{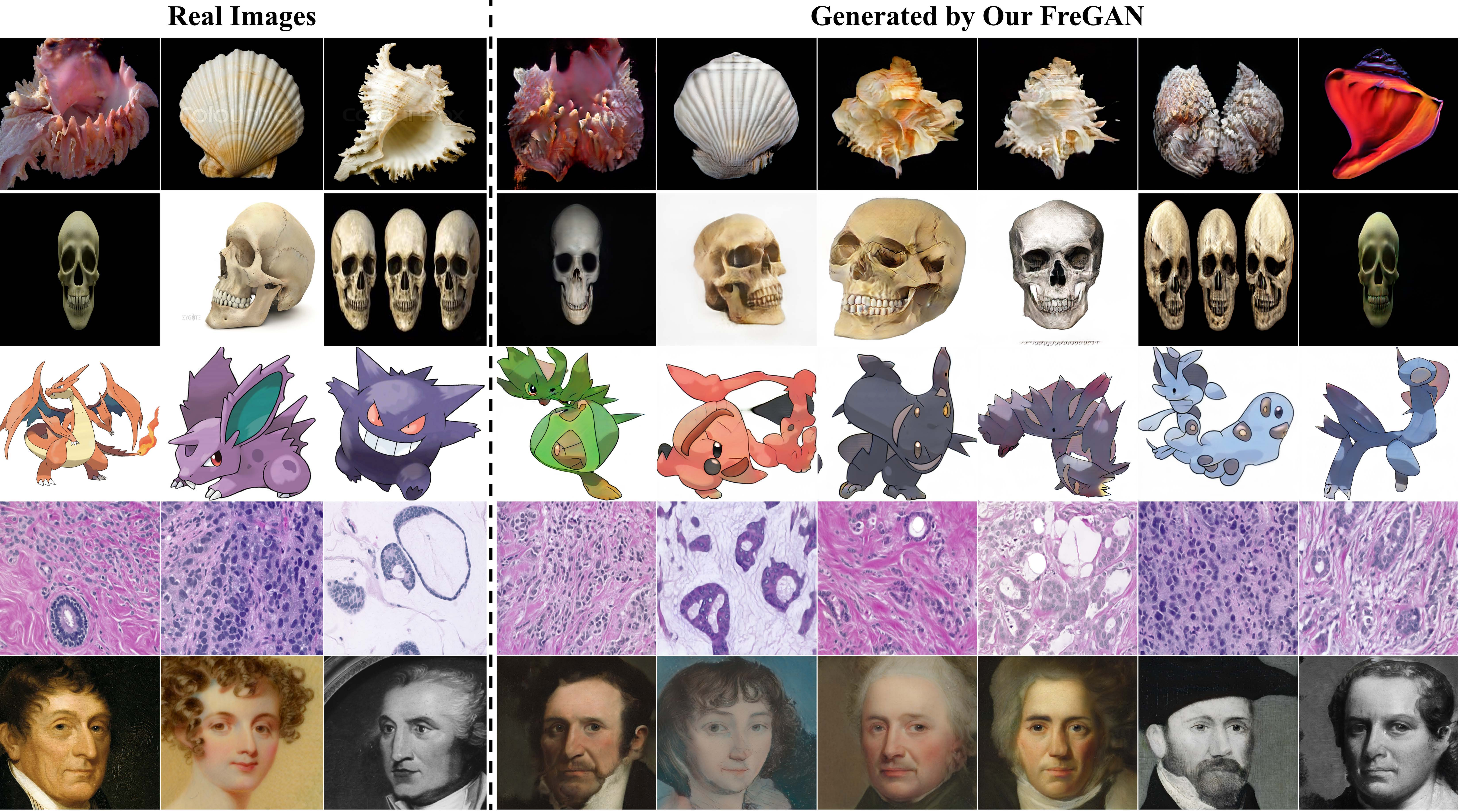}
	\vspace{-0.4cm}
    \caption{\textbf{Qualitative results of our FreGAN.} The left part shows some of real training images and the right part of images are generated by our FreGAN.
    Our FreGAN is capable of generating photorealistic images with fine details, which is indistinguishable by the discriminator.}
	\label{fig:Generated1024Appendix}
	\vspace{-3mm}
\end{figure}

We give more generated images of our FreGAN in Fig.~\ref{fig:Generated256Appendix}, Fig.~\ref{fig:Generated512Appendix}, and Fig.~\ref{fig:Generated1024Appendix}.
We can observe from these figures that our FreGAN can produce vivid images with fine details.
The photorealistic generated images indicate the effectiveness of our FreGAN in improving the generation quality of GANs under limited data.
However,  our FreGAN still struggles in generating photorealistic images when given datasets with limited data but various contents, \eg, only dozens of images, and their contents vary widely.
As can be seen from Fig.~\ref{fig:Generated1024Appendix}, the overall quality of the generated Pokemon images is not satisfactory because the Pokemon dataset contains multiple categories of Pokemon and
only a few images for each character, making it hard to produce high-quality Pokemon images.


\begin{figure}
	\centering
	\vspace{-0.1cm}
	\includegraphics[width=\linewidth]{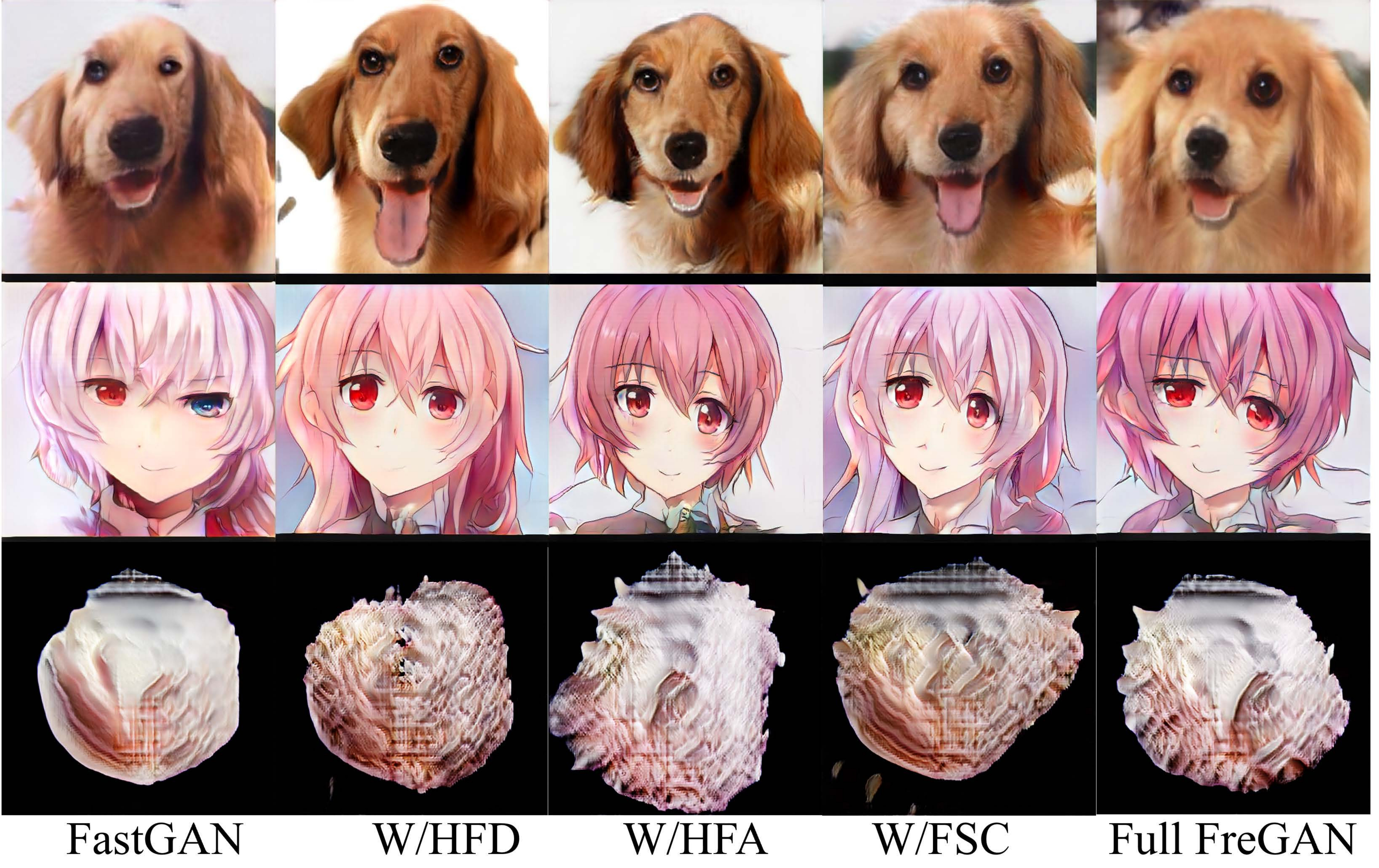}
	\vspace{-0.4cm}
    \caption{{\textbf{Qualitative comparison results of ablation studies.} Each component of our method contributes to the quality of the generated images. HFD improves the fine details like the eyes of the dog and the color of eyes and hairs of the AnimeFace, and HFA makes nostrils, teeth more realistic.}}
	\label{fig:AblationVis}
	\vspace{-3mm}
\end{figure}

\newpage

\section{Analysis on generation diversity}
\label{sec:diversityapp}
\begin{figure}
	\centering
	\vspace{-0.1cm}
	\includegraphics[width=\linewidth]{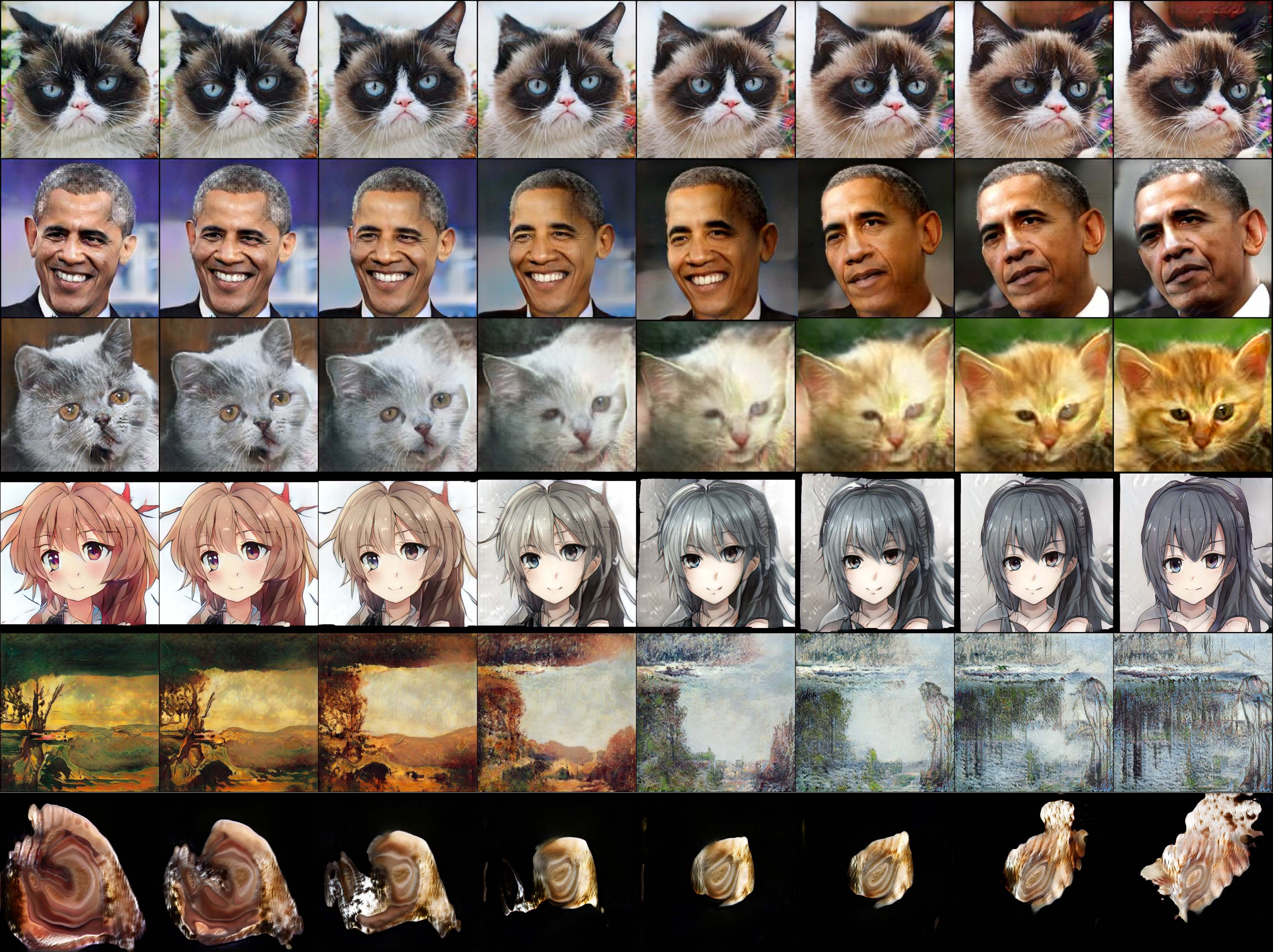}
	\vspace{-0.4cm}
    \caption{{\textbf{Latent space interpolation results of our FreGAN.} The smooth transition images suggest that our FreGAN is capable of generating images instead of memorizing images.}}
	\label{fig:Interpolation}
	\vspace{-3mm}
\end{figure}

\begin{figure}
	\centering
	\vspace{-0.1cm}
	\includegraphics[width=\linewidth]{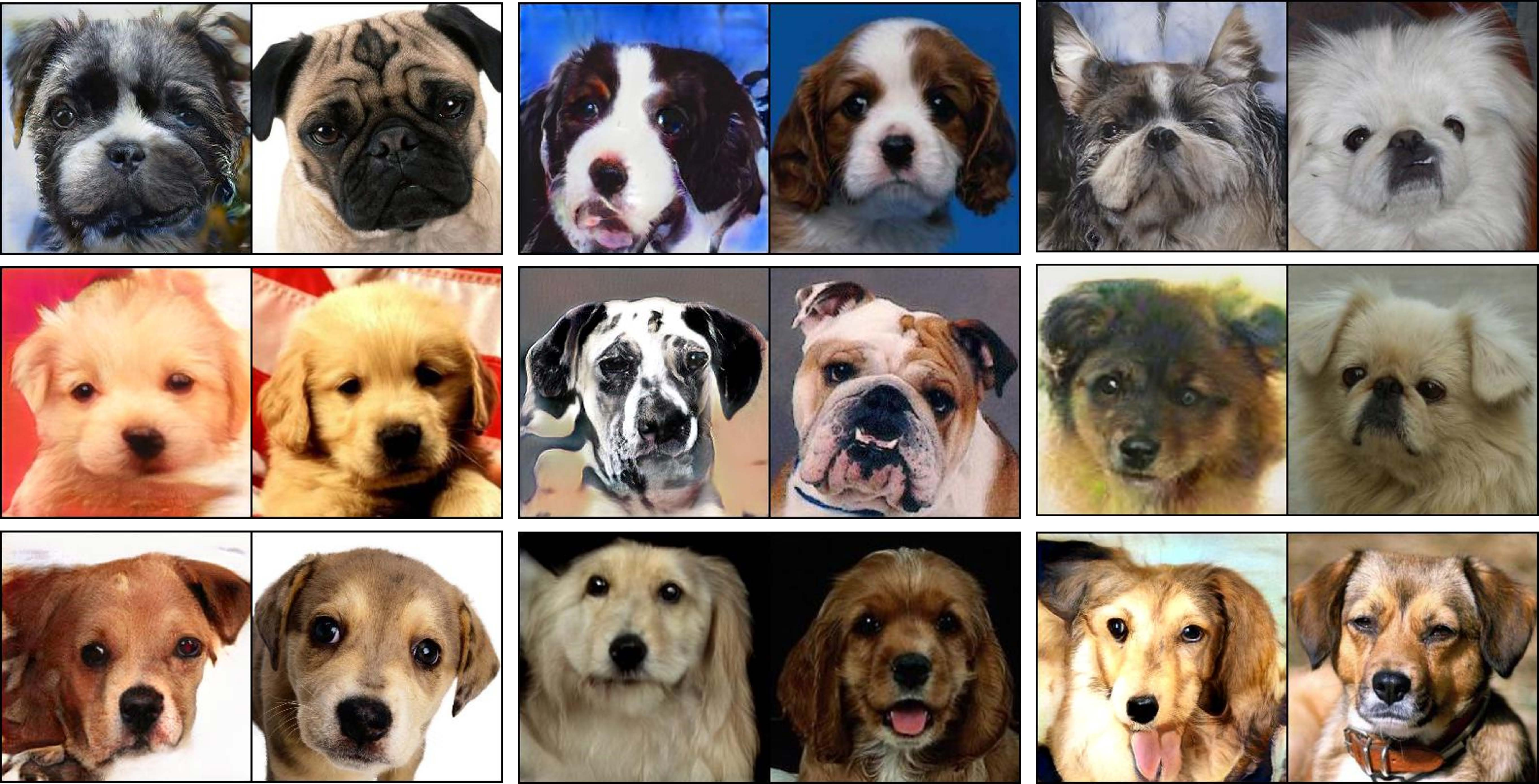}
	\vspace{-0.4cm}
    \caption{{\textbf{Nearest real samples to the generated ones on AnimalFace Dog dataset.} For each paired images, the left one is generated by our FreGAN and the right one is the closest image found from the training data. We adopt LPIPS score to measure the similarity of images, \emph{i.e.}, the closest image found from the training data have the largest LPIPS score with the generated one.}}
	\label{fig:NN256}
	\vspace{-3mm}
\end{figure}

\begin{figure}
	\centering
	\vspace{-0.1cm}
	\includegraphics[width=\linewidth]{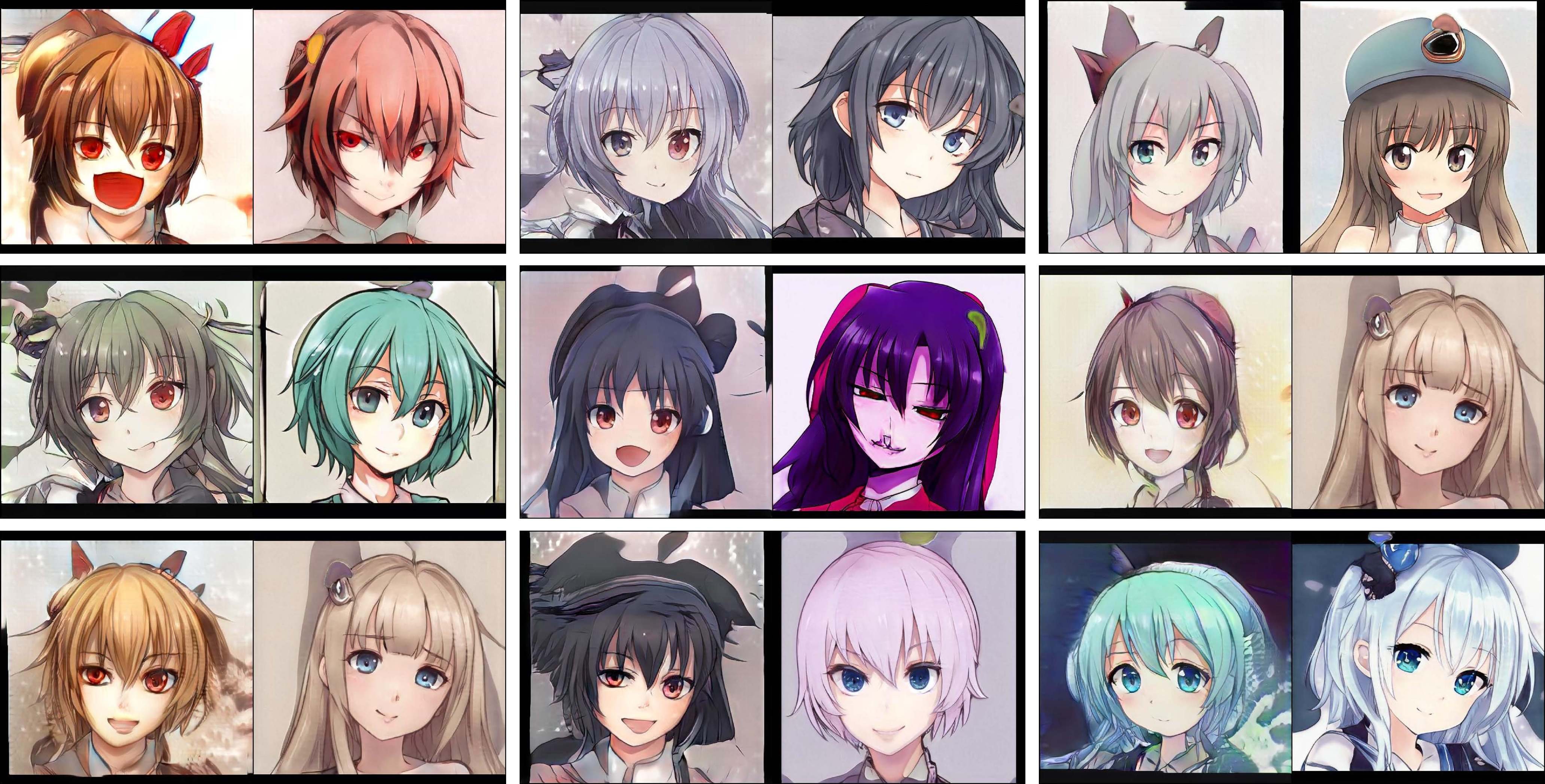}
	\vspace{-0.4cm}
     \caption{{\textbf{Nearest real samples to the generated ones on AnimeFace dataset.} For each paired images, the left one is generated by our FreGAN and the right one is the closest image found from the training data. We adopt LPIPS score to measure the similarity of images, \emph{i.e.}, the closest image found from the training data have the largest LPIPS score with the generated one.}}
    \label{fig:NN512}
	\vspace{-3mm}
\end{figure}

\begin{figure}
	\centering
	\vspace{-0.1cm}
	\includegraphics[width=\linewidth]{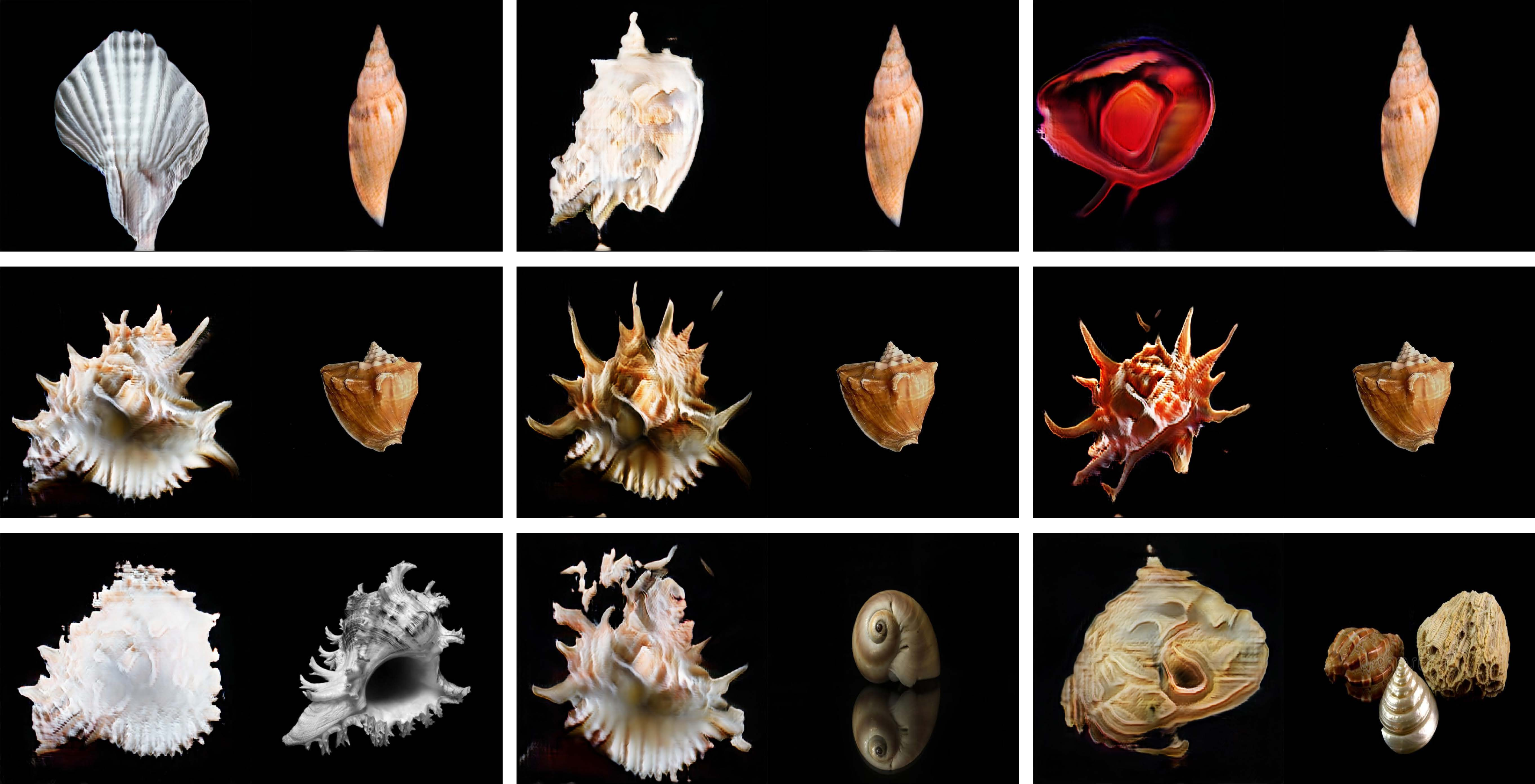}
	\vspace{-0.4cm}
    \caption{{\textbf{Nearest real samples to the generated ones on Shells dataset.} For each paired images, the left one is generated by our FreGAN and the right one is the closest image found from the training data. We adopt LPIPS score to measure the similarity of images, \emph{i.e.}, the closest image found from the training data have the largest LPIPS score with the generated one.}}
	\label{fig:NN1024}
	\vspace{-3mm}
\end{figure}

\begin{figure}
	\centering
	\vspace{-0.1cm}
	\includegraphics[width=\linewidth]{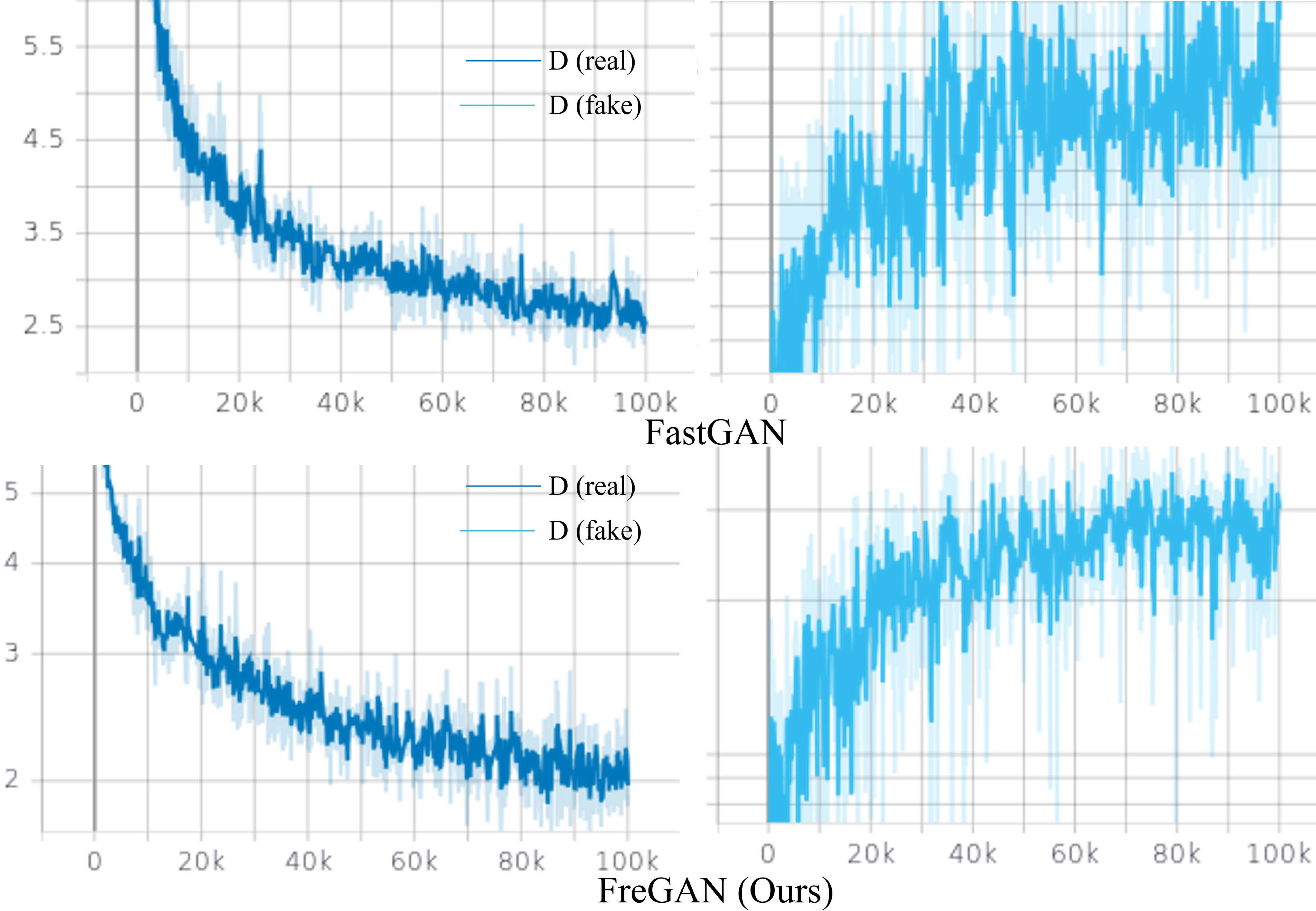}
	\vspace{-0.4cm}
    \caption{{\textbf{Comparison results of the outputs of the discriminator throughout the training.} The outputs of the discriminator on fake images of our FreGAN rise more stable and continuous, indicating that our generator is trained more effectively, leading to better generation quality.}}
	\label{fig:LossD}
	\vspace{-3mm}
\end{figure}

{We provide the latent space interpolation results of our FreGAN in Fig.~\ref{fig:Interpolation}, from which we can observe that the transition of images generated by different latent codes are smooth and photo-realistic, indicating that our FreGAN promotes the generation quality without compromising the generation diversity.
Moreover, we find the closest real images to the generated ones from training data based on LPIPS score, the visualization results are given in Fig.~\ref{fig:NN256}, Fig.~\ref{fig:NN512}, and Fig.~\ref{fig:NN1024}. The results demonstrate that our FreGAN learns to produce new images instead of memorizing training images. For example, the body hair color, perspective, and demeanor of dogs are different. The mouth, eyes, hairstyle of AnimeFace are different. And for the Shells dataset in Fig.~\ref{fig:NN1024}, different generated images that have the closest distance with the same real image are different in color, shape, etc, which further demonstrating that our method improves generation quality meanwhile maintaining diversity.
To investigate the training process of our FreGAN, we plot the outputs of the discriminator throughout the training in Fig.~\ref{fig:LossD}, the stable curve of our FreGAN demonstrate that our FreGAN can be trained more effective.}

\newpage

\section{More analysis on spectral properties of generated images}
\label{sec:spectralanaapp}
\begin{figure}
	\centering
	\vspace{-0.1cm}
	\includegraphics[width=\linewidth]{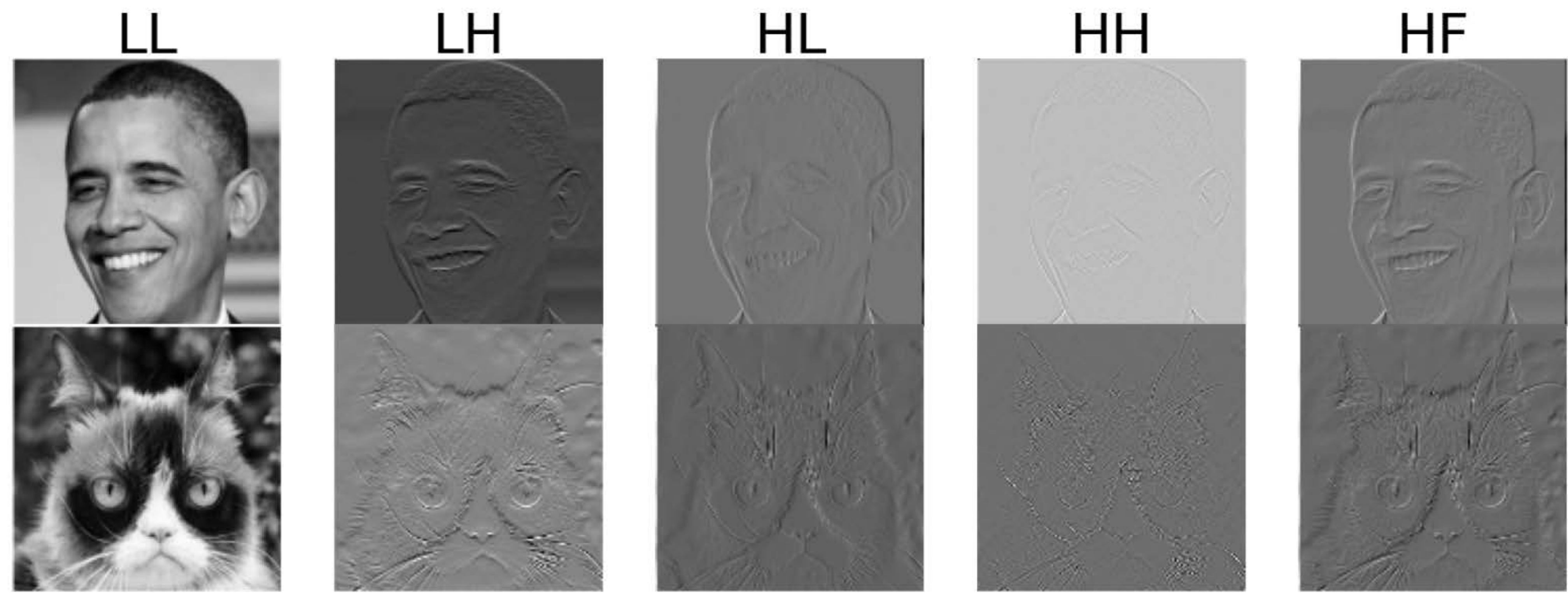}
	\vspace{-0.4cm}
    \caption{{\textbf{2D DWT illustration results of differen frequency components.} The low (L) pass filter captures images' overall textures and outlines, and the high (H) pass filter concentrates on details such as the background and edges.}}
	\label{fig:2DDWTIntro}
	\vspace{-3mm}
\end{figure}

\begin{figure}
	\centering
	\vspace{-0.1cm}
	\includegraphics[width=\linewidth]{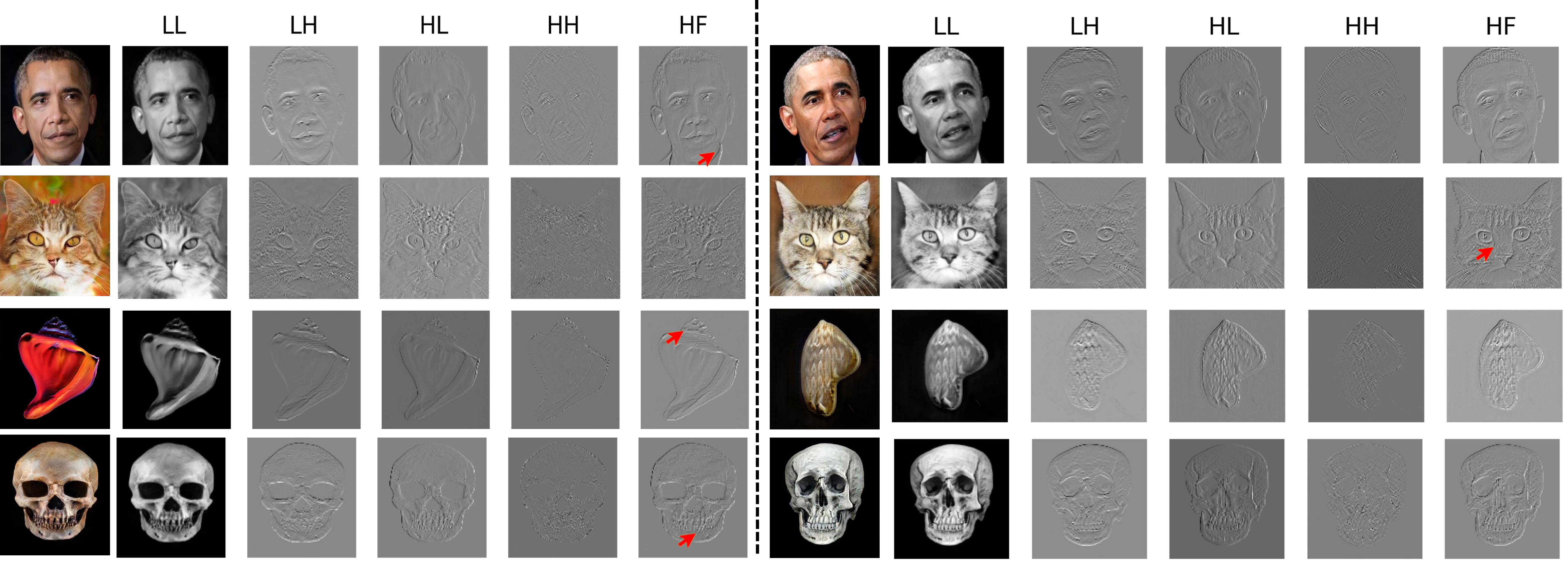}
	\vspace{-0.4cm}
    \caption{{\textbf{2D DWT Qualitative comparison results of our FreGAN and the baseline FastGAN.} The images from left to right are generated images, 2D DWT LL, LH, HL, HH, and the combined High-frequency components respectively.
    Our FreGAN improves the overall quality of generated images and raises the model's frequency awareness, encouraging the generator to produce precise high-frequency signals with fine details.}}
	\label{fig:2DDWT1}
	\vspace{-3mm}
\end{figure}

\begin{figure}
	\centering
	\vspace{-0.1cm}
	\includegraphics[width=\linewidth]{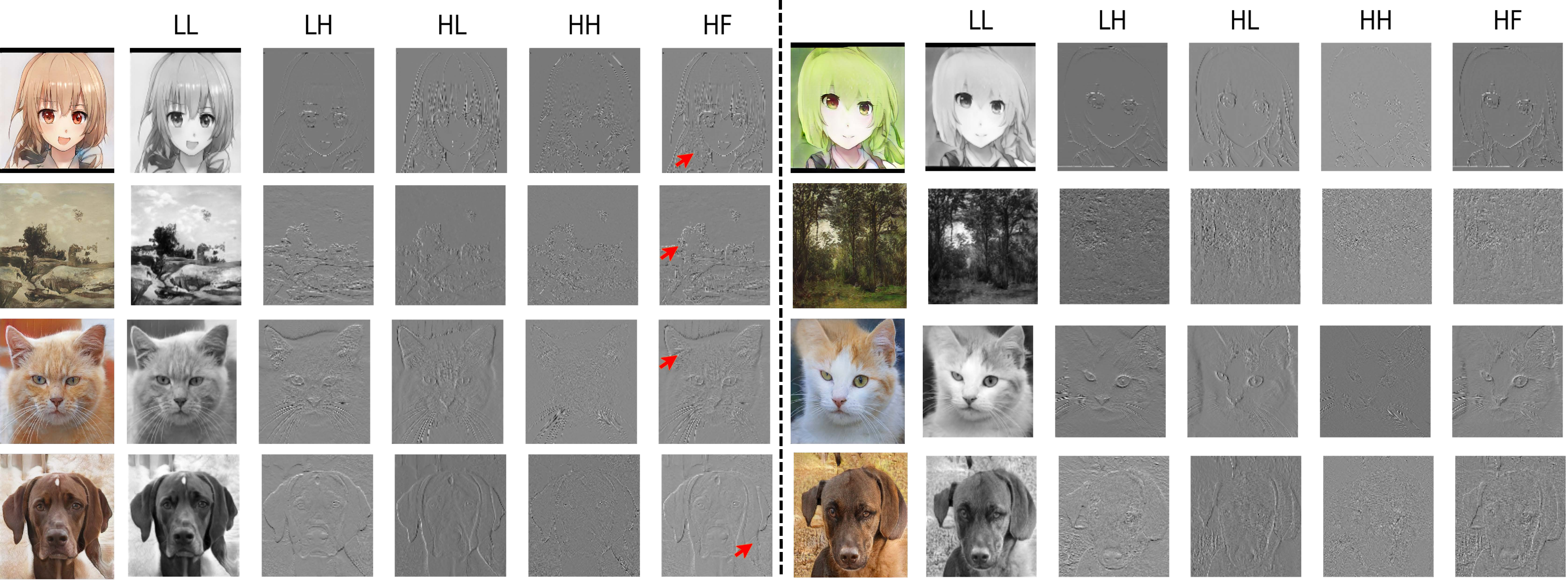}
	\vspace{-0.4cm}
    \caption{{\textbf{2D DWT Qualitative comparison results of our FreGAN and the baseline FastGAN.} The images from left to right are generated images, 2D DWT LL, LH, HL, HH, and the combined High-frequency components respectively.
    Our FreGAN improves the overall quality of generated images and raises the model's frequency awareness, encouraging the generator to produce precise high-frequency signals with fine details.}}
	\label{fig:2DDWT2}
	\vspace{-3mm}
\end{figure}

\begin{figure}
	\centering
	\vspace{-0.1cm}
	\includegraphics[width=\linewidth]{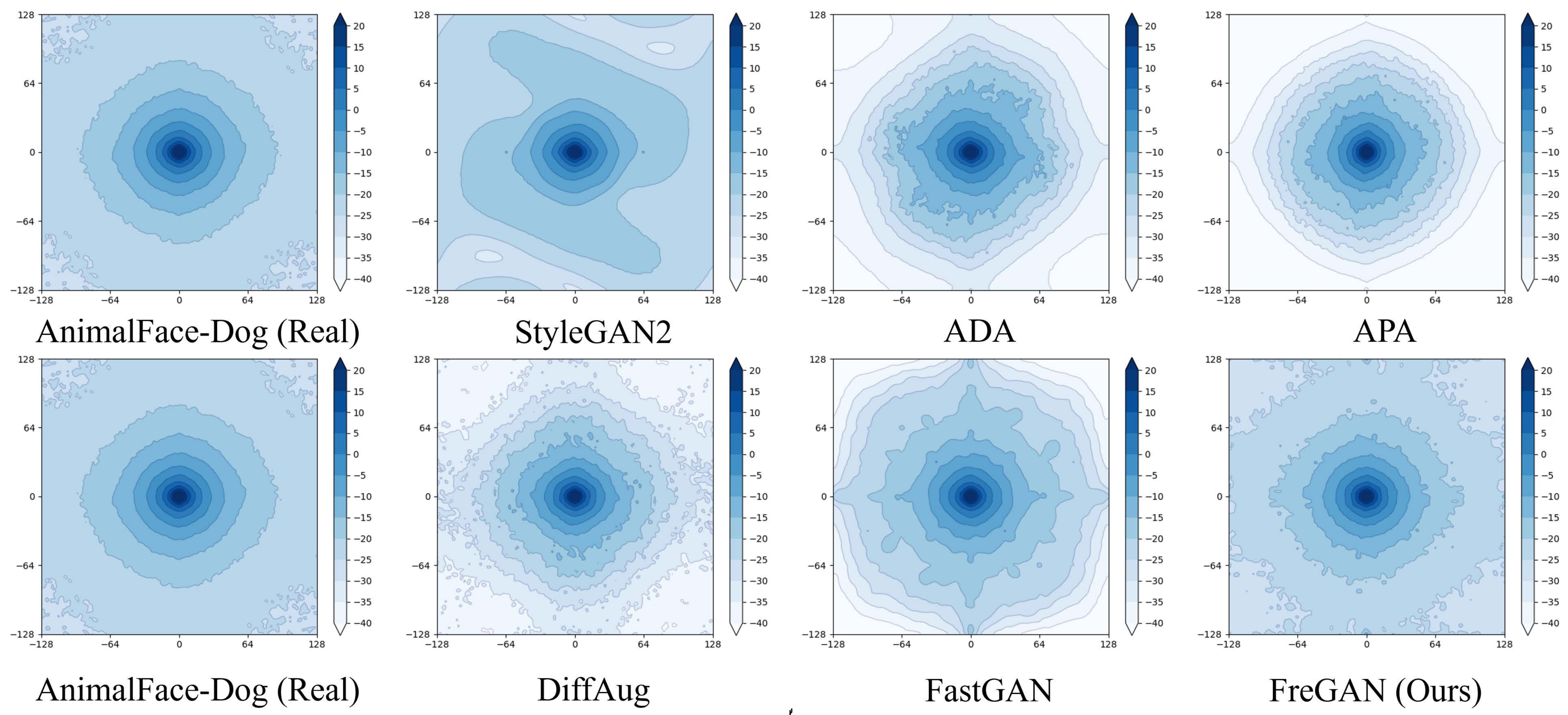}
	\vspace{-0.4cm}
    \caption{{\textbf{Comparison results of average 2D power spectrum~\cite{Karras2021} on Animal Face Dog.} The average 2D power spectrum result for the real data is computed from all training data, and the results of our FreGAN and compared methods are computed from 5k generated images.}}
	\label{fig:FrequencyHeatmap}
	\vspace{-3mm}
\end{figure}

\begin{figure}
	\centering
	\vspace{-0.1cm}
	\includegraphics[width=\linewidth]{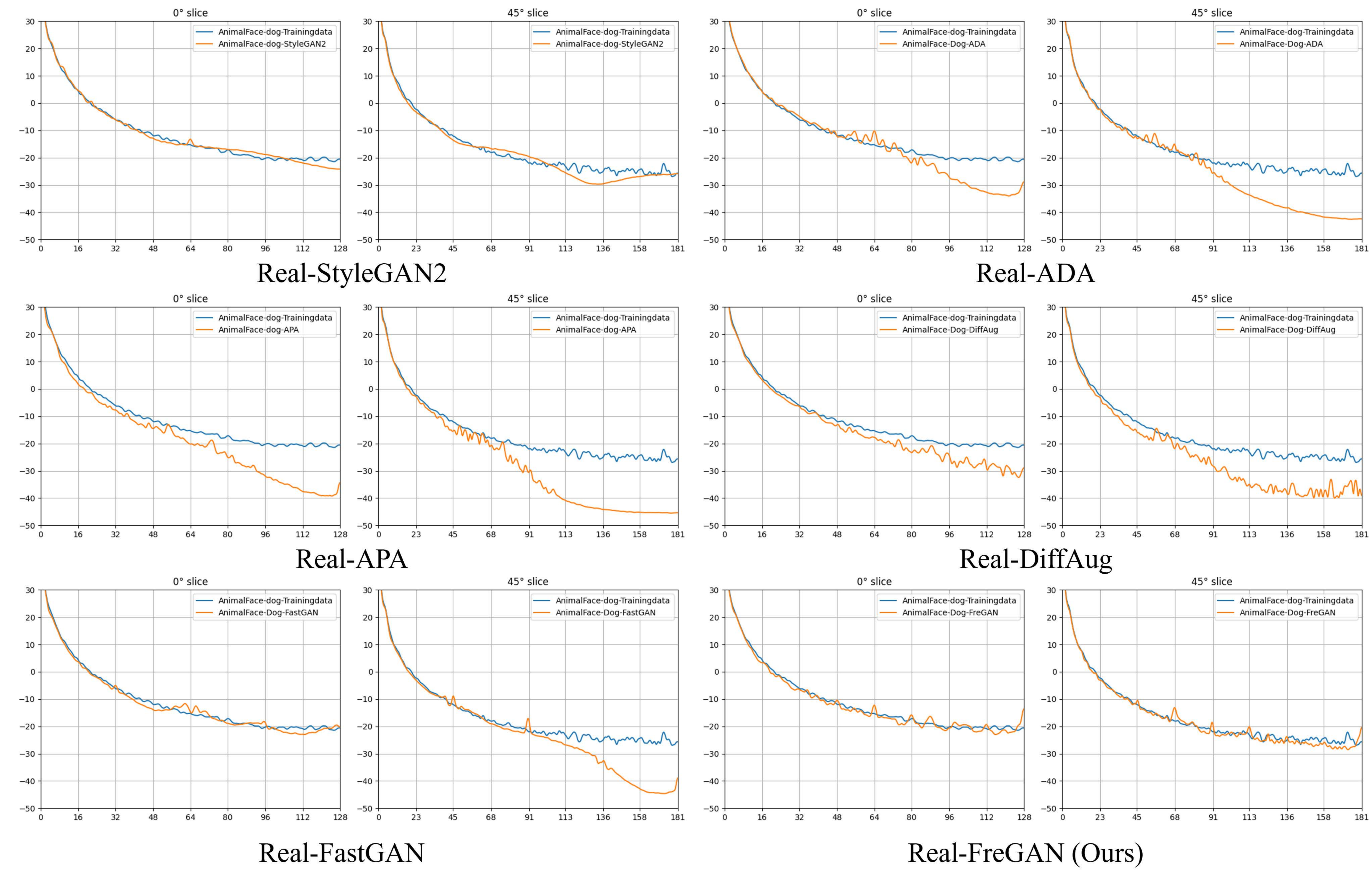}
	\vspace{-0.4cm}
    \caption{{\textbf{Comparison results of one-dimensional slices of the power spectrum~\cite{Karras2021} on Animal Face Dog.} The 1D slices of the spectrum are computed along the horizontal angle (0\degree) without azimuthal integration~\cite{Karras2021}.}}
	\label{fig:FrequencySlice}
	\vspace{-3mm}
\end{figure}

\begin{figure}
	\centering
	\vspace{-0.1cm}
	\includegraphics[width=\linewidth]{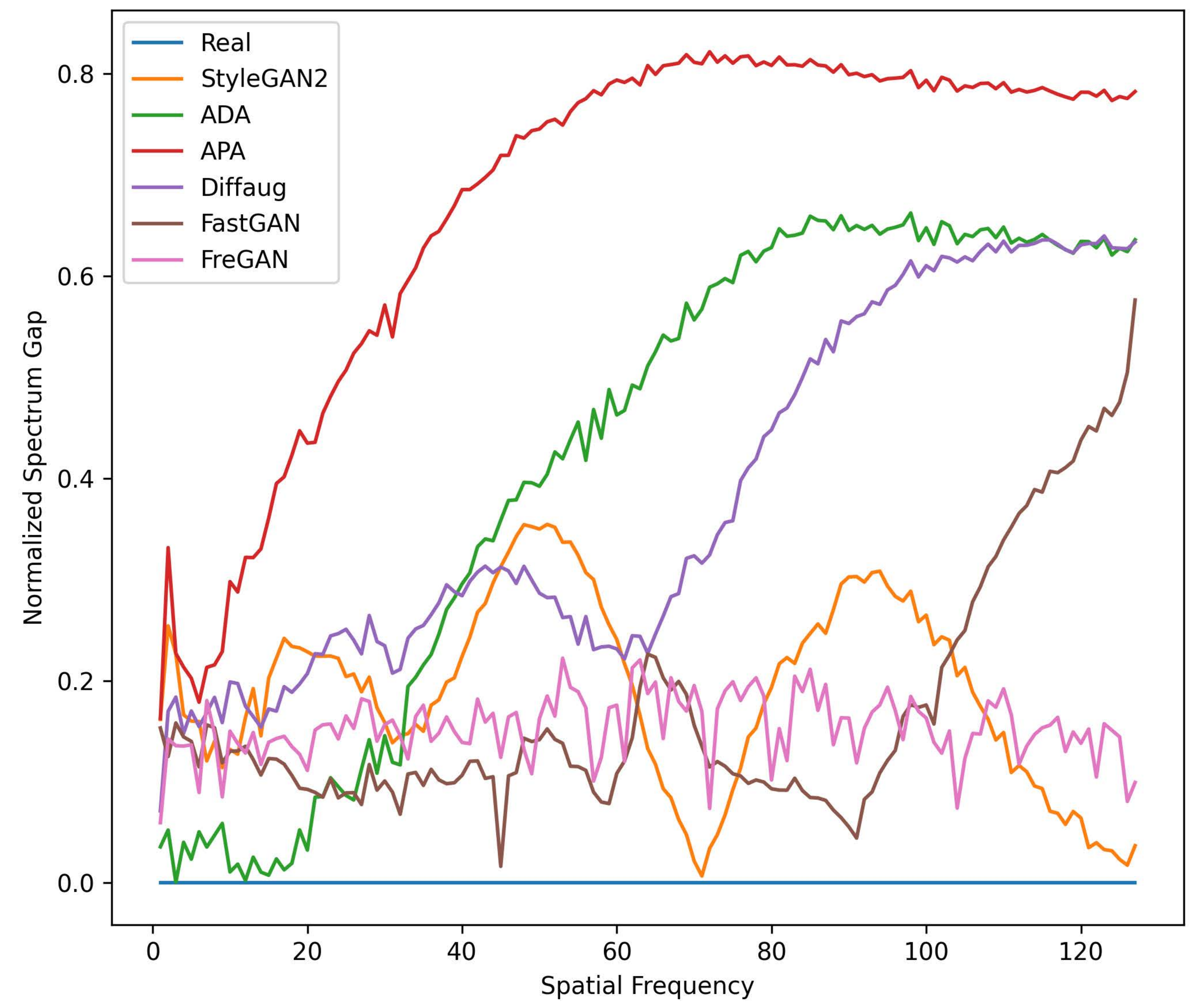}
	\vspace{-0.4cm}
    \caption{{\textbf{The power spectrum distance~\cite{gal2021swagan} between real images and those generated by our FreGAN and compared methods.} The gap between the images generated by our FreGAN and the real images is the smallest, demonstrating that our FreGAN can produce more realistic frequency signals, especially high-frequency signals as shown in the right half of the figure.}}
	\label{fig:powerspectrumgap}
	\vspace{-3mm}
\end{figure}

\begin{figure}
	\centering
	\vspace{-0.1cm}
	\includegraphics[width=\linewidth]{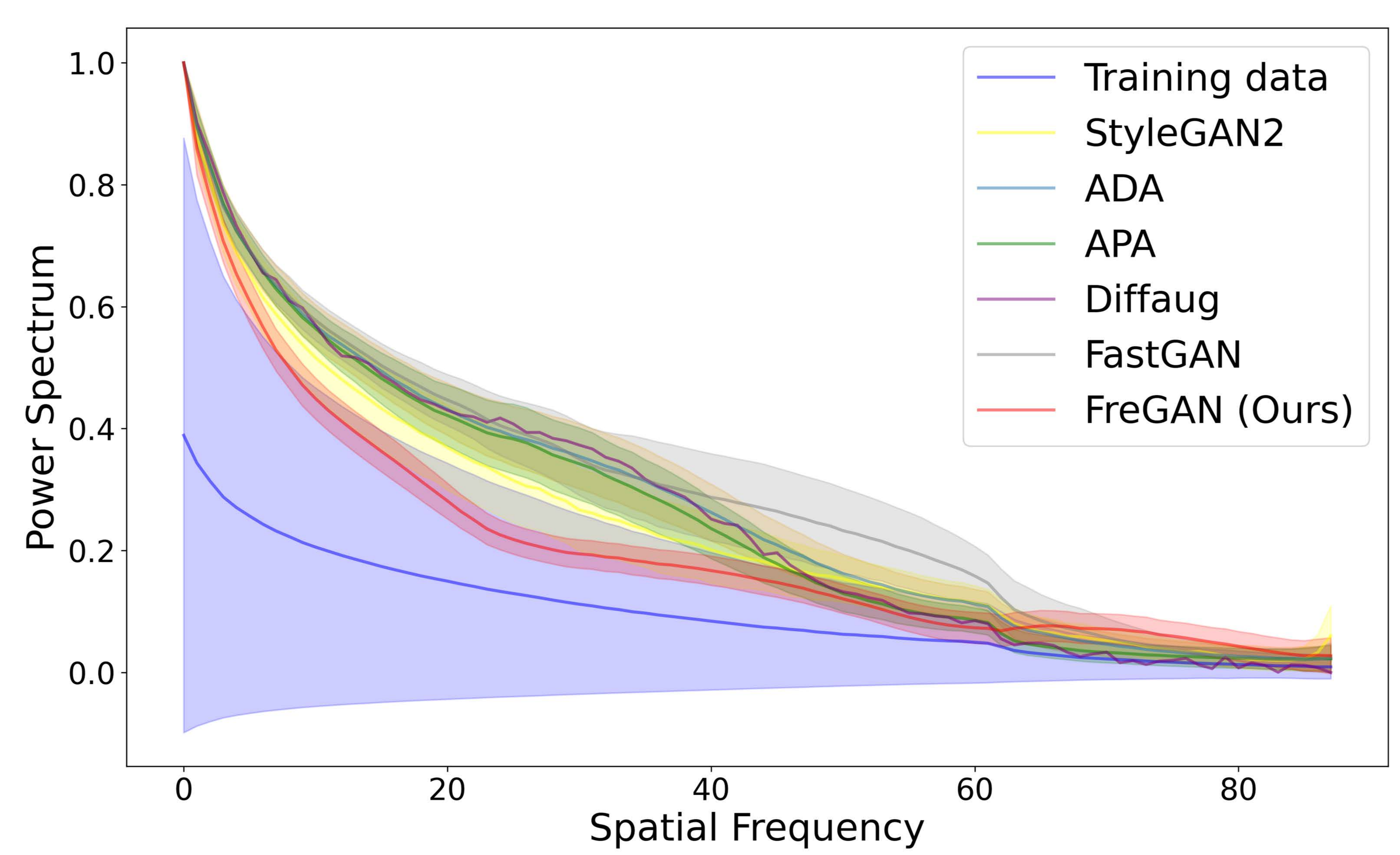}
	\vspace{-0.4cm}
    \caption{{\textbf{Power Spectrum~\cite{durall2020watch} comparison results of our FreGAN and baselines on Animal Face Dog dataset.} The statistics (mean and variance) results are obtained after azimuthal integration over the power-spectrum. The statistics of our method overlap the most with that of the training data, indicating that our model is frequency-aware and can produce more realistic frequency signals.}}
	\label{fig:PowerSpectrum}
	\vspace{-3mm}
\end{figure}

{We first give 2D DWT results of real images in Fig.~\ref{fig:2DDWTIntro}.
Then we provide the 2D DWT visualization results in Fig.~\ref{fig:2DDWT1} and Fig.~\ref{fig:2DDWT2}, which complement the qualitative comparison results in Figure.4 of the main paper.
We can observe that although presented in different ways of visualization, images generated by our FreGAN contain more realistic frequency signals, indicating the efficacy of our proposed techniques.
Besides, we compare the averaged 2D power spectrum, one-dimensional slices of the power spectrum, the power spectrum distance, and the statistic (mean and variance power spectrum) in Fig.~\ref{fig:FrequencyHeatmap}, Fig.~\ref{fig:FrequencySlice}, Fig.~\ref{fig:powerspectrumgap}, and Fig.~\ref{fig:PowerSpectrum}, respectively.
We can observe from these figures that:
1) Our FreGAN can produce more realistic frequency signals compared with other methods;
2) The overlap between the generated images of our FreGAN and the training data is the largest (Fig.~\ref{fig:PowerSpectrum});
3) Our FreGAN is stable and can consistently produce effective frequency signals (Fig.~\ref{fig:FrequencySlice} and Fig.~\ref{fig:powerspectrumgap});}

\newpage

\end{document}